\pdfoutput=1

\documentclass[11pt]{article}

\usepackage{acl}

\usepackage{times}
\usepackage{latexsym}

\usepackage[T1]{fontenc}

\usepackage[utf8]{inputenc}

\usepackage{microtype}

\usepackage{xspace}
\usepackage{hyperref}
\usepackage[normalem]{ulem}
\usepackage{algorithm}
\usepackage{algpseudocode}
\usepackage{mathtools}
\usepackage{amssymb}
\usepackage{booktabs, multirow}
\usepackage{makecell}
\usepackage{subcaption}
\usepackage{graphicx}
\usepackage{rotating}
\usepackage{tikz}

\def\mPLM{mPLM\xspace}
\def\mPLMs{mPLMs\xspace}
\def\mPLMMeasure{mPLM-Sim\xspace}
\def\Bible{PBC\xspace}
\def\Flores{Flores\xspace}
\def\Fleurs{Fleurs\xspace}

\title{\mPLMMeasure:
Better Cross-Lingual Similarity and Transfer in Multilingual Pretrained Language Models}

\author{Peiqin Lin$^*$$^{1,2}$, Chengzhi Hu$^*$$^{3,7}$, Zheyu Zhang$^{1}$, André F. T. Martins$^{4,5,6}$, Hinrich Schütze$^{1,2}$ \\
        $^1$Center for Information and Language Processing, LMU Munich \\
        $^2$Munich Center for Machine Learning \quad
        $^3$Institute of Informatics, LMU Munich \\
        $^4$Instituto Superior Técnico, Universidade de Lisboa (Lisbon ELLIS Unit) \quad
        $^5$Unbabel \\
        $^6$Instituto de Telecomunicações  \quad
        $^7$Konrad Zuse School of Excellence in Reliable AI\\
        \texttt{linpq@cis.lmu.de, \{Chengzhi.Hu, Zheyu.Zhang\}@campus.lmu.de}
}

\newcounter{notecounter}
\newcommand{\enotesoff}{\long\gdef\enote##1##2{}}

\enotesoff

\begin{document}
\maketitle

\def\thefootnote{*}\footnotetext{Equal contribution.}\def\thefootnote{\arabic{footnote}}

\begin{abstract}
Recent multilingual pretrained language models (\mPLMs) have been shown to encode strong language-specific signals, which are not explicitly provided during pretraining.
It remains an open question whether it is feasible to employ \mPLMs to measure language similarity, and subsequently use the similarity results to select source languages for boosting cross-lingual transfer.
To investigate this, we propose \mPLMMeasure, a language
similarity measure that induces the similarities across languages from \mPLMs using multi-parallel corpora. Our study shows that \mPLMMeasure exhibits moderately high correlations with linguistic similarity measures, such as lexicostatistics, genealogical language family, and geographical sprachbund. We also conduct a case study on languages with low correlation and observe that \mPLMMeasure yields more accurate similarity results. Additionally, we find that similarity results vary across different mPLMs and different layers within an mPLM.
We further investigate whether \mPLMMeasure is effective for zero-shot cross-lingual transfer by conducting experiments on both low-level syntactic tasks and high-level semantic tasks. The experimental results demonstrate that \mPLMMeasure is capable of selecting better source languages than linguistic measures, resulting in a 1\%-2\% improvement in zero-shot cross-lingual transfer performance.\footnote{Our code is open-sourced at \url{https://github.com/cisnlp/mPLM-Sim}.}
\end{abstract}

\section{Introduction}
\label{sec:intro}

Recent multilingual pretrained language models (\mPLMs) trained with massive data, e.g., mBERT \citep{DBLP:conf/naacl/DevlinCLT19}, XLM-R \citep{DBLP:conf/acl/ConneauKGCWGGOZ20} and BLOOM \citep{DBLP:journals/corr/abs-2211-05100}, have become a standard for multilingual representation learning. Follow-up works \citep{DBLP:conf/emnlp/WuD19,DBLP:conf/emnlp/LibovickyRF20,DBLP:journals/corr/abs-2109-08040,DBLP:journals/corr/abs-2205-10964} show that these \mPLMs encode strong language-specific signals which are not explicitly provided during pretraining.
However, the possibility of using \mPLMs to measure language similarity and utilizing the similarity results to pick source languages for enhancing cross-lingual transfer is not yet thoroughly investigated.

To investigate language similarity in \mPLMs, we
propose \mPLMMeasure, a measure that leverages \mPLMs and multi-parallel corpora to measure similarity between languages. Using \mPLMMeasure, we intend to answer the following research questions.

\textbf{(Q1)} \textit{What is the correlation between \mPLMMeasure and linguistic similarity?}

We compute Pearson correlation between similarity results of \mPLMMeasure and linguistic similarity measures. The results show that \mPLMMeasure has a moderately high correlation with some linguistic measures, such as lexical-based and language-family-based measures. Additional case studies on languages with low correlation demonstrate that \mPLMs can acquire the similarity patterns among languages through pretraining on massive data.

\textbf{(Q2)} \textit{Do different layers of an \mPLM produce different similarity results?}

\citet{DBLP:conf/acl/JawaharSS19,DBLP:conf/emnlp/SabetDYS20,DBLP:journals/coling/ChoenniS22}
have demonstrated that different linguistic information is
encoded across different layers of an \mPLM. We analyze the
performance of \mPLMMeasure across layers and show
that \mPLMMeasure results vary across layers, aligning with
previous findings. Specifically, the embedding layer
captures lexical information, whereas the middle layers
reveal more intricate similarity patterns encompassing
general, geographical, and syntactic aspects. However, 
in the high layers,
the ability to distinguish
between languages becomes less prominent. Furthermore, we
observe that clustering of languages also varies by layer,
shedding new light on how the representation of
language-specific information changes throughout layers.

\textbf{(Q3)} \textit{Do different \mPLMs produce different similarity results?}

We make a comprehensive comparison among a diverse set of 11 \mPLMs in terms of architecture, modality, model size, and tokenizer. The experimental results show that input modality (text or speech), model size, and data used for pretraining have large effects on \mPLMMeasure while tokenizers and training objectives have little effect.

\textbf{(Q4)} \textit{Can \mPLMMeasure choose better source languages for zero-shot cross-lingual transfer?}

Previous works 
\citep{DBLP:conf/acl/LinCLLZXRHZMALN19,DBLP:conf/acl/PiresSG19,DBLP:conf/emnlp/LauscherRVG20,DBLP:journals/corr/abs-2212-09651,DBLP:journals/corr/abs-2305-00090,DBLP:conf/acl/ImaiKOO23}
have shown that the performance of cross-lingual transfer
positively correlates with linguistic similarity. However,
we find that there can be a mismatch between \mPLM subspaces
and linguistic clusters, which may lead to a failure of
zero-shot cross-lingual transfer for low-resource
languages. Intuitively, \mPLMMeasure can select the source
languages that boost cross-lingual transfer better than
linguistic similarity since it captures the subspaces
learned during pretraining (and which are the basis for successful transfer). To examine this, we conduct experiments on four datasets that require reasoning about different levels of syntax and semantics for a diverse set of low-resource languages. The results show that \mPLMMeasure achieves 1\%-2\% improvement over linguistic similarity measures for cross-lingual transfer.

\section{Setup}
\label{sec:setup}

\subsection{\mPLMMeasure}
\label{sec:method}

Generally, a transformer-based \mPLM consists of $N$ layers:
$N - 1$ transformer layers plus the static
embedding layer. Given a multi-parallel
corpus\footnote{Monolingual corpora covering multiple
languages can be also used to measure language
similarity. Our initial experiments
(\S\ref{sec:cmp_corpora1}) show that parallel corpora yield
better results while using fewer sentences than monolingual
corpora. Therefore, we use parallel corpora for our investigation.}, \mPLMMeasure
aims to provide the similarity results of $N$ layers for
an \mPLM across $L$ languages considered. In this context,
we define languages using the ISO 639-3 code combined with
the script, e.g., ``eng\_Latn''
represents English written in Latin.

For each sentence $x$ in the multi-parallel corpus, the \mPLM computes its sentence embedding for the $i$th layer of the \mPLM: $\boldsymbol{h}_i = E(x)$. For \mPLMs with bidirectional encoders, including encoder architecture, e.g., XLM-R, and encoder-decoder architecture, e.g., mT5, $E(\cdot)$ is a mean pooling operation over hidden states, which performs better than [CLS] and MAX strategies \citep{DBLP:conf/emnlp/ReimersG19}. For \mPLMs with auto-regressive encoders, e.g., mGPT, $E(\cdot)$ is a position-weighted mean pooling method, which gives later tokens a higher weight \citep{DBLP:journals/corr/abs-2202-08904}. Finally, sentence embeddings for all sentences of the $L$ languages are obtained.

For $i$th layer, the similarity of each language pair is computed using the sentence embeddings of all multi-parallel sentences. Specifically, we get the cosine similarity of each parallel sentence of the language pair, and then average all similarity scores across sentences as the final score of the pair. Finally, we have a similarity matrix $\boldsymbol{S}_i \in \mathbb{R}^{L \times L}$ across $L$ languages for the $i$th layer of the \mPLM.

\begin{table*}
    \centering
    \resizebox{\textwidth}{!}{
    \begin{tabular}{c|cccccccc}
    \toprule
    Model & Size & |Lang| & |Layer| & Tokenizer & Arch. & Objective & Modality & Data \\ \midrule
    mBERT \citep{DBLP:conf/naacl/DevlinCLT19} & 172M & 104 & 13 & Subword & Enc & MLM, NSP & Text & Wikipedia \\
    XLM-R-Base \citep{DBLP:conf/acl/ConneauKGCWGGOZ20} & 270M & 100 & 13 & Subword & Enc & MLM & Text & CC \\
    XLM-R-Large \citep{DBLP:conf/acl/ConneauKGCWGGOZ20} & 559M & 100 & 25 & Subword & Enc & MLM & Text & CC \\
    Glot500 \citep{imanigooghari2023glot500} & 395M & 515 & 13 & Subword & Enc & MLM & Text & Glot500-c \\
    mGPT \citep{DBLP:journals/corr/abs-2204-07580} & 1.3B & 60 & 25 & Subword & Dec & CLM & Text & Wikipedia+mC4 \\
    mT5-Base \citep{DBLP:conf/naacl/XueCRKASBR21} & 580M & 101 & 13 & Subword & Enc-Dec & MLM & Text & mC4 \\
    CANINE-S \citep{DBLP:journals/tacl/ClarkGTW22} & 127M & 104 & 17 & Char & Enc & MLM, NSP & Text & Wikipedia \\
    CANINE-C \citep{DBLP:journals/tacl/ClarkGTW22} & 127M & 104 & 17 & Char & Enc & MLM, NSP & Text & Wikipedia \\
    XLM-Align \citep{DBLP:conf/acl/Chi0ZHMHW20} & 270M & 94 & 13 & Subword & Enc & MLM, TLM, DWA & Text & Wikipedia+CC \\
    NLLB-200 \citep{DBLP:journals/corr/abs-2207-04672} & 1.3B & 204 & 25 & Subword & Enc-Dec & MT & Text & NLLB \\
    XLS-R-300M \citep{DBLP:journals/corr/abs-2111-09296} & 300M & 128 & 25 & - & Enc & MSP & Speech & CommonVoice \\
    \bottomrule
    \end{tabular}
    }
    \caption{11 \mPLMs considered in the paper. |Layer| denotes the number of layers used for measuring similarity. Both the static embedding layer and all layers of the transformer are considered. For encoder-decoder architectures, we only consider the encoder. |Lang|: the number of languages covered. Arch.: Architecture. Enc: Encoder. Dec: Decoder. MLM: Masked Language Modeling. CLM: Causal Language Modeling. TLM: Translation Language Modeling. NSP: Next Sentence Prediction. DWA:  Denoising Word Alignment. MT: Machine Translation. MSP: Masked Speech Prediction. CC: CommonCrawl.}
    \label{tab:model}
\end{table*}

\subsection{\mPLMs, Corpora and Languages}

We consider a varied set of 11 \mPLMs for our investigation, differing in model size, number of covered languages, architecture, modality, and data used for pretraining. Full list and detailed information of the selected \mPLMs are shown in Tab.~\ref{tab:model}.

We work with three multi-parallel corpora: the text
corpora \Flores \citep{DBLP:journals/corr/abs-2207-04672}
and Parallel Bible Corpus (\Bible, \citep{DBLP:conf/lrec/MayerC14}) and the speech
corpus
\Fleurs \citep{DBLP:journals/corr/abs-2205-12446}. \Flores
covers more than 200 languages. Since both \Bible
and \Fleurs are not fully multi-parallel, we reconstruct
them to make them multi-parallel. After
recostruction, \Bible covers 379 languages, while \Fleurs
covers 67 languages. \Bible consists of religious text, and
both \Flores and \Fleurs are from web articles. The speech
of \Fleurs is aligned to the text of \Flores, enabling
us to compare text \mPLMs with speech \mPLMs. We use 500
multi-parallel sentences from each corpus. Languages covered
by \mPLMs and corpora are listed in \S\ref{sec:languages}.

\begin{table*}
    \centering
    \resizebox{\textwidth}{!}{
    \begin{tabular}{cccccccc}
    \toprule
    Task & Corpus & |Train| & |Dev| & |Test| & |Lang| & Metric & Domain \\
    \midrule
    \multirow{2}{*}{\makecell{Sequence \\ Labeling}} & NER \citep{DBLP:conf/acl/PanZMNKJ17} & 5,000 & 500 & 100-10,000 & 108 & F1 & Wikipedia \\
    & POS \citep{DBLP:journals/coling/MarneffeMNZ21} & 5,000 & 500 & 100-22,358 & 60 & F1 & Misc \\
    \bottomrule
    \multirow{2}{*}{\makecell{Text \\ Classification}} & MASSIVE \citep{DBLP:journals/corr/abs-2204-08582} & 11,514 & 2,033 & 2,974 & 44 & Acc & Misc \\
    & Taxi1500 \citep{ma2023taxi1500} & 860 & 106 & 111 & 130 & F1 & Bible \\
    \bottomrule
    \end{tabular}
    }
    \caption{Evaluation dataset
    statistics.
    |Train|/|Dev|: train/dev set
    size (source language).
    |Test|: test set size (target language).
    |Lang|: number of target languages.}
    \label{tab:tasks}
\end{table*}

\subsection{Evaluation}

\paragraph{Pearson Correlation}

We compute Pearson correlation scores to measure how
much \mPLMMeasure correlates with seven linguistic
similarity measures:
LEX, GEN, GEO, SYN, INV, PHO and FEA.
LEX is computed
based on the edit distance of the two corpora. The
six others
are provided by lang2vec. GEN is based on
language family. GEO is orthodromic distance,
i.e.,
the shortest distance between two points on the surface of
the earth.
SYN is derived from the syntactic
structures of the languages. Both INV and PHO are
phonological features. INV is derived from PHOIBLE, while
PHO is based on WALS and Ethnologue. FEA is
computed by combining GEN, GEO, SYN, INV and PHO.

For each target language, we have the similarity scores
between the target language and the other $L-1$ languages
based on the similarity matrix $\boldsymbol{S}_i$ for layer
$i$ (see \S\ref{sec:method}), and also the similarity scores based on the
considered linguistic similarity measure $j$.
Then we compute the Pearson correlation $r_{i}^{j}$ between these two similarity score lists. We choose the highest correlation score across all layers as the result of each target language since the results for different languages vary across layers. Finally, we report MEAN (M) and MEDIAN (Mdn) of the correlation scores for all languages. Here, we consider 32 languages covered by all models and corpora.

\paragraph{Case Study} In addition to the quantitative evaluation, we conduct manual analysis for languages that exhibit low correlation scores. We apply complete linkage hierarchical clustering to get the similar languages of the analyzed language for analysis. Specifically, the languages which have the most common shared path in the hierarchical tree with the target language are considered as similar languages. To analyze as many languages as possible, we consider the setting of Glot500 and \Bible.

\begin{table*}
    \centering
    \resizebox{\textwidth}{!}{
    \begin{tabular}{c|cc|cc|cc|cc|cc|cc}
        \toprule
        & \multicolumn{2}{c|}{XLM-R-Base} & \multicolumn{2}{c|}{XLM-R-Large} & \multicolumn{2}{c|}{mT5-Base} & \multicolumn{2}{c|}{mGPT} & \multicolumn{2}{c|}{mBERT} & \multicolumn{2}{c}{Glot500} \\ \cline{2-13}
        & M & Mdn & M & Mdn & M & Mdn & M & Mdn & M & Mdn & M & Mdn \\ \midrule
        LEX & 0.740 & 0.859 & 0.684 & 0.862 & 0.628 & 0.796 & 0.646 & 0.848 & 0.684 & 0.882 & 0.741 & 0.864 \\
        GEN & 0.489 & 0.563 & 0.570 & 0.609 & 0.577 & 0.635 & 0.415 & 0.446 & 0.513 & 0.593 & 0.527 & 0.600 \\
        GEO & 0.560 & 0.656 & 0.587 & 0.684 & 0.528 & 0.586 & 0.348 & 0.362 & 0.458 & 0.535 & 0.608 & 0.674 \\
        SYN & 0.637 & 0.662 & 0.709 & 0.738 & 0.594 & 0.612 & 0.548 & 0.591 & 0.611 & 0.632 & 0.577 & 0.607 \\
        INV & 0.272 & 0.315 & 0.312 & 0.292 & 0.295 & 0.321 & 0.340 & 0.394 & 0.216 & 0.246 & 0.248 & 0.293 \\
        PHO & 0.112 & 0.151 & 0.207 & 0.258 & 0.166 & 0.176 & 0.184 & 0.239 & 0.111 & 0.125 & 0.094 & 0.144 \\
        FEA & 0.378 & 0.408 & 0.443 & 0.466 & 0.354 & 0.371 & 0.455 & 0.479 & 0.346 & 0.361 & 0.358 & 0.372 \\ \midrule
        AVG & 0.455 & 0.516 & 0.502 & 0.559 & 0.449 & 0.500 & 0.420 & 0.480 & 0.420 & 0.482 & 0.451 & 0.508 \\ \midrule
        & \multicolumn{2}{c|}{CANINE-S} & \multicolumn{2}{c|}{CANINE-C} & \multicolumn{2}{c|}{NLLB-200} & \multicolumn{2}{c|}{XLM-Align} & \multicolumn{2}{c|}{XLS-R-300M} & \multicolumn{2}{c}{AVG} \\ \cline{2-13}
        & M & Mdn & M & Mdn & M & Mdn & M & Mdn & M & Mdn & M & Mdn \\ \midrule
        LEX & 0.661 & 0.821 & 0.639 & 0.784 & 0.722 & 0.856 & 0.728 & 0.869 & 0.285 & 0.262 & 0.651 & 0.791 \\
        GEN & 0.548 & 0.629 & 0.565 & 0.633 & 0.538 & 0.626 & 0.516 & 0.606 & 0.401 & 0.353 & 0.514 & 0.572 \\
        GEO & 0.504 & 0.560 & 0.533 & 0.624 & 0.490 & 0.499 & 0.616 & 0.690 & 0.531 & 0.541 & 0.524 & 0.583 \\
        SYN & 0.476 & 0.521 & 0.507 & 0.559 & 0.375 & 0.370 & 0.634 & 0.669 & 0.354 & 0.389 & 0.548 & 0.577 \\
        INV & 0.329 & 0.390 & 0.369 & 0.406 & 0.337 & 0.373 & 0.252 & 0.315 & 0.191 & 0.180 & 0.287 & 0.321 \\
        PHO & 0.112 & 0.137 & 0.117 & 0.173 & 0.101 & 0.108 & 0.105 & 0.143 & 0.124 & 0.115 & 0.130 & 0.161 \\
        FEA & 0.317 & 0.297 & 0.367 & 0.360 & 0.311 & 0.326 & 0.368 & 0.399 & 0.203 & 0.175 & 0.355 & 0.365 \\ \midrule
        AVG & 0.421 & 0.479 & 0.442 & 0.506 & 0.411 & 0.451 & 0.460 & 0.527 & 0.298 & 0.288 & 0.430 & 0.481 \\
        \bottomrule
    \end{tabular}
    }
    \caption{Comparison across \mPLMs: Pearson correlation between \mPLMMeasure and seven similarity measures for all \mPLMs and \Flores/\Fleurs on 32 languages. \mPLMMeasure strongly correlates with LEX, moderate strongly correlates with GEN, GEO, and SYN, and weakly correlates with INV, PHO, and FEA.}
    \label{tab:overall}
\end{table*}

\paragraph{Cross-Lingual Transfer} To compare \mPLMMeasure
with linguistic measures for zero-shot cross-lingual
transfer, we run experiments for low-resource languages on four datasets, including two for sequence labeling, and two for text classification. Details of the four tasks are shown in Tab.~\ref{tab:tasks}.

We selected six high-resource and typologically diverse languages, namely Arabic (arb\_Arab), Chinese (cmn\_Hani), English (eng\_Latn), Hindi (hin\_Deva), Russian (rus\_Cyrl), and Spanish (spa\_Latn), as source languages.
For a fair comparison, we use the same amount of source language data for fine-tuning and validation as shown in Tab.~\ref{tab:tasks}.

The evaluation targets all languages that are covered by both Glot500 and \Flores and have at least 100 samples, excluding the six source languages.
The language list for evaluation is provided in \S\ref{sec:languages}.

We obtain the most similar source language for each target
language by applying each of the seven linguistic similarity
measures
(LEX, GEN, GEO, SYN, INV, PHO, FEA)
and our \mPLMMeasure. Here, we consider the setting of
Glot500 and \Flores for \mPLMMeasure since extensive
experiments (see \S\ref{sec:cmp_corpora2}) show that \Flores
provides slightly better similarity results than \Bible.
For the linguistic similarity measures,
if
the most similar source language is not available due to
missing values in lang2vec, we use eng\_Latn as the source
language. We also compare \mPLMMeasure with the ENG
baseline defined as using eng\_Latn as the source language for all target languages.

We use the same hyper-parameter settings as
in \citep{DBLP:conf/icml/HuRSNFJ20,DBLP:journals/corr/abs-2204-08582,ma2023taxi1500}. Specifically,
we set the batch size to 32 and the learning rate to 2e-5
for both NER and POS, and fine-tune Glot500 for 10 epochs. For
MASSIVE, we use a batch size of 16, a learning rate of
4.7e-6, and train for 100 epochs. For Taxi1500, we use a
batch size of 32, a learning rate of 2e-5, and train for 30
epochs. In all tasks, we select the model for evaluating
target languages based on the performance of the source
language validation set.

\section{Results}

\subsection{Comparison Between \mPLMMeasure and Linguistic Similarity}
\label{sec:correlation}

Tab.~\ref{tab:overall} shows the Pearson correlation between \mPLMMeasure and linguistic similarity measures of 11 \mPLMs, and also the average correlations of all 11 \mPLMs. We observe that \mPLMMeasure strongly correlates with LEX, which is expected since \mPLMs learn language relationships from data and LEX similarity is the easiest pattern to learn. Besides, \mPLMMeasure has moderately strong correlations with GEN, GEO, and SYN, which shows that \mPLMs can learn high-level patterns for language similarity. \mPLMMeasure also has a weak correlation with INV, and a very weak correlation with PHO, indicating \mPLMs do not capture  phonological similarity well. Finally, \mPLMMeasure correlates with FEA weakly since FEA is the measure combining both high- and low-correlated linguistics features.

To further compare \mPLMMeasure with linguistic similarity measures, we conduct a manual analysis on languages for which \mPLMMeasure has weak correlations with LEX, GEN, and GEO. As mentioned in \S\ref{sec:setup}, with the setting of Glot500 and \Bible, we apply hierarchical clustering and use similar results for analysis.

We find that \mPLMMeasure can deal well with languages that are not covered by lang2vec. For example, Norwegian Nynorsk (nno\_Latn) is not covered by lang2vec, and \mPLMMeasure can correctly find its similar languages, 
i.e., Norwegian Bokmål (nob\_Latn) and Norwegian (nor\_Latn).
Furthermore, \mPLMMeasure can well capture the similarity between languages which cannot be well measured by either LEX, GEN, or GEO.

For LEX, \mPLMMeasure can capture similar languages written
in different scripts. A special case is the same languages
in different scripts. Specifically, \mPLMMeasure matches
Uighur in Latin and Arabic (uig\_Arab and uig\_Latn), also
Karakalpak in Latin and Cyrillic (kaa\_Latn and
kaa\_Cyrl). In general, \mPLMMeasure does a good job at clustering languages from the same language family but written in different scripts, e.g., Turkic (Latn, Cyrl, Arab) and Slavic (Latn, Cyrl).

For GEN, \mPLMMeasure captures correct similar languages for
isolates and constructed languages. Papantla Totonac
(top\_Latn) is a language of the Totonacan language family
and spoken in Mexico. It shares areal features with the Nahuan
languages (nch\_Latn, ncj\_Latn, and ngu\_Latn) of
the Uto-Aztecan family, which are all located in the
Mesoamerican language
area.\footnote{\url{https://en.wikipedia.org/wiki/Mesoamerican_language_area}}
Esperanto (epo\_Latn) is a constructed language whose
vocabulary derives primarily from Romance languages,
and \mPLMMeasure correctly identifies
Romance languages such as French
(fra\_Latn) and Italian (ita\_Latn) as similar. The above two cases
show the superiority of \mPLMMeasure compared to GEN.

The GEO measure may not be suitable for certain language families, such as Austronesian languages and mixed languages. Austronesian languages have the largest geographical span among language families prior to the spread of Indo-European during the colonial period.\footnote{\url{https://en.wikipedia.org/wiki/Austronesian_languages}} Moreover, for mixed languages, such as creole languages, their similar languages are often geographically distant due to colonial history. In contrast to GEO, \mPLMMeasure can better cluster these languages.

The above analysis shows that it is non-trivial to use either LEX, GEN, or GEO for measuring language similarity. In contrast, \mPLMMeasure directly captures similarity from \mPLMs and can therefore produce better similarity results.

However, we observe that obtaining accurate similarity
results from \mPLMs using \mPLMMeasure can be challenging
for certain languages. To gain further insights into this
issue, we examine the correlation between performances,
specifically the correlation between \mPLMMeasure and GEN,
and the sizes of the pretraining data. Surprisingly, we find
a remarkably weak correlation (-0.008), suggesting that
differences in pretraining data sizes do not significantly
contribute to variations in performances.

Instead, our findings indicate a different key factor: the coverage of multiple languages within the same language family. This observation is substantiated by a strong correlation of 0.617 between the diversity of languages within a language family (measured by the number of languages included) and the performance of languages belonging to that particular language family.

\subsection{Comparison Across Layers for \mPLMMeasure}
\label{cmp_layers}

\begin{figure}
  \centering
  \resizebox {\columnwidth} {!} {
    \includegraphics[trim={0.6cm 0.6cm 0.6cm 0.6cm},clip]{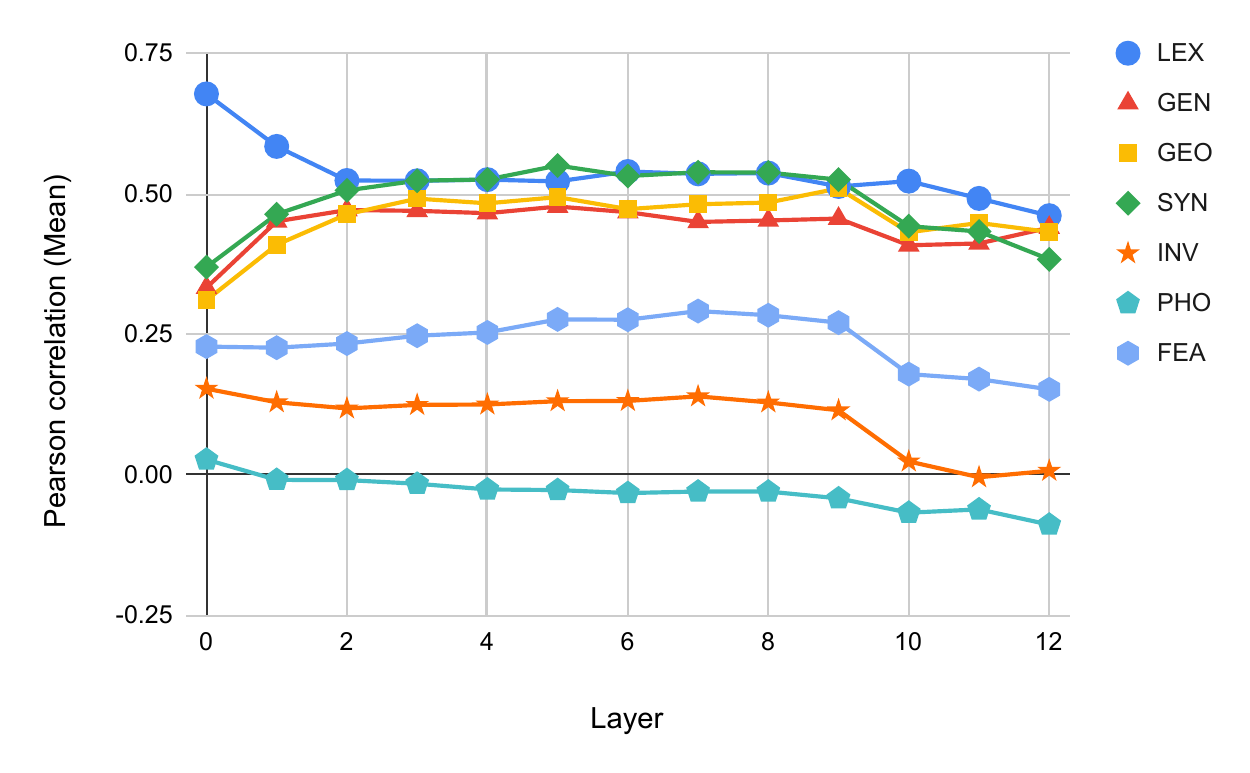}
  }
  \caption{Comparison across layers: Pearson correlation (MEAN) between \mPLMMeasure and linguistic similarity measures across layers for Glot500 and \Flores on 32 languages. Correlation between \mPLMMeasure and LEX peaks in the first layer and decreases, while the correlation with GEN, GEO, and SYN slightly increases in the low layers before reaching its peak.}
  \label{fig:layer}
\end{figure}

\begin{figure*}
  \centering
  \resizebox {\textwidth} {!} {
    \includegraphics[trim={4cm 0cm 4cm 0cm},clip]{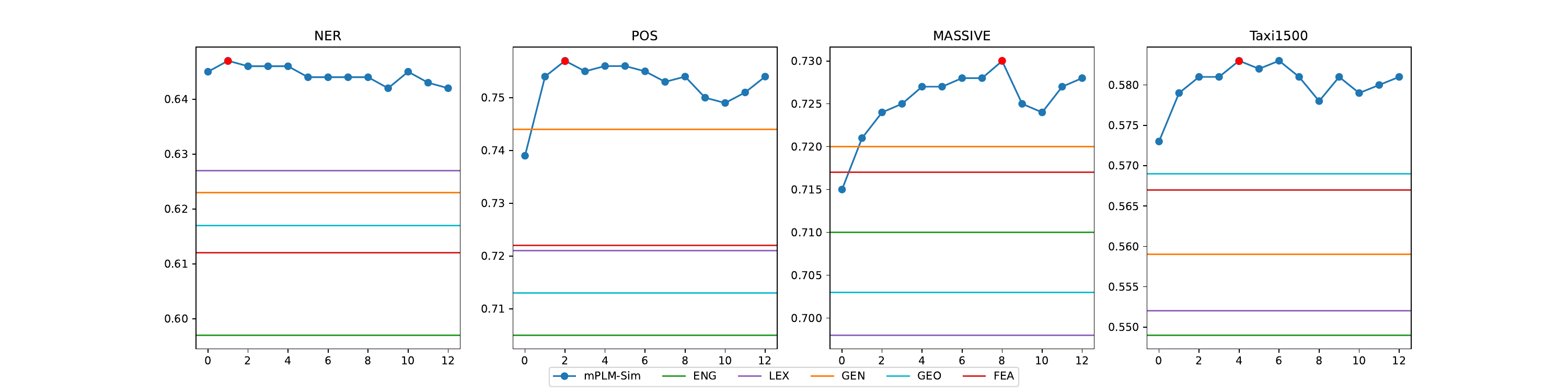}
  }
  \caption{Macro average results
(averaged over
target languages) on cross-lingual transfer
for baselines and for
\mPLMMeasure in all layers of Glot500. ENG represents using English as the source language. LEX, GEN, GEO, and FEA indicate using the most similar languages based on the corresponding similarity measures as the source language. The red dots of \mPLMMeasure highlight the layer with the highest score.
  }
  \label{fig:cross-lingual result}
\end{figure*}

We analyze the correlation between \mPLMMeasure and linguistic similarity measures across different layers of an \mPLM, specifically for Glot500. The results, presented in Fig.~\ref{fig:layer}, demonstrate the variation in \mPLMMeasure results across layers. Notably, in the first layer, \mPLMMeasure exhibits a high correlation with LEX, which gradually decreases as we move to higher layers. Conversely, the correlation between \mPLMMeasure and GEN, GEO, and SYN shows a slight increase in the lower layers, reaching its peak in layer 1 or 2 of the \mPLM. However, for the higher layers (layers 10-12), all correlations slightly decrease. We also performed further visualization and analysis  across layers
using the setting of Glot500 and \Flores
for \mPLMMeasure (\S\ref{sec:va_layer}). The findings are consistent with our observations from Fig.~\ref{fig:layer}.

Furthermore, our case study shows that the layers which have highest correlations between \mPLMMeasure and LEX, GEN, or GEO vary across languages. For example, Atlantic–Congo languages achieve highest correlation with GEN at the 1st layer, while Mayan languages at the 6th layer. This finding demonstrates that language-specific information changes across layers.

\subsection{Comparison Across Models for \mPLMMeasure}
\label{sec:cmp_model}

Tab.~\ref{tab:overall} presents a broad comparison among 11 different \mPLMs, revealing several key findings.

Firstly, the decoder architecture has a negative impact on performance due to the inherent difficulty in obtaining accurate sentence-level representations from the decoder. For example, the decoder-only \mPLM mGPT performs worse than encoder-only \mPLMs such as XLM-R and mBERT. This observation is reinforced by the comparison between XLM-R-Large and mT5-Base, which have nearly identical model sizes. Remarkably, XLM-R-Large outperforms mT5-Base on AVG by 5\% for both Mean (M) and Median (Mdn) scores. 

Additionally, tokenizer-free \mPLMs achieve comparable performance to subword-tokenizer-based \mPLMs. Notably, \mPLMs such as mBERT, CANINE-S, and CANINE-C, which share pretraining settings, exhibit similar performances.

The size of \mPLMs also influences \mPLMMeasure in terms of LEX, GEN, and SYN. Comparing XLM-R-Base with XLM-R-Large, higher-level language similarity patterns are more evident in larger \mPLMs. Specifically, XLM-R-Large shows a higher correlation with high-level patterns such as GEN and SYN, while having a lower correlation with low-level patterns like LEX, compared to XLM-R-Base.

The training objectives adopted in \mPLMs also impact the performance of \mPLMMeasure. Task-specific \mPLMs, such as NLLB-200, perform slightly worse than general-purpose \mPLMs. Besides, XLM-Align, which leverages parallel objectives to align representations across languages, achieves comparable results to XLM-R-Base. This highlights the importance of advancing methods to effectively leverage parallel corpora.

The choice of pretraining data is another important factor. For example, mBERT uses Wikipedia, while XLM-R-Base uses CommonCrawl, which contains more code-switching. As a result, XLM-R-Base has a higher correlation with GEO and achieves higher AVG compared to mBERT.

The speech \mPLM, i.e., XLS-R-300M, exhibits lower correlation than text \mPLMs, consistent with findings from \citet{abdullah2023nature}. XLS-R-300M learns language similarity from speech data, which is biased towards the accents of speakers. Consequently, XLS-R-300M has a higher correlation with GEO, which is more related to accents, than other similarity measures.

Factors such as the number of languages have minimal effects on \mPLMMeasure. Glot500, covering over 500 languages, achieves comparable results with XLM-R-Base.

\subsection{Effect for Cross-Lingual Transfer}
\label{sec:cross-lingual transfer}

\begin{table*}
    \centering
    \resizebox{\textwidth}{!}{
    \begin{tabular}{c|c|c|cc|cc|c|c|c|cc|cc|c}
    \toprule
    &  & Language & \multicolumn{2}{c|}{GEN} & \multicolumn{2}{c|}{mPLM-Sim} & $\Delta$ &  & Language & \multicolumn{2}{c|}{GEN} & \multicolumn{2}{c|}{mPLM-Sim} & $\Delta$ \\
    \midrule
    \midrule
    \multirow{3}{*}{\rotatebox[origin=c]{90}{high end}} & \multirow{6}{*}{\rotatebox[origin=c]{90}{NER}} & jpn\_Jpan & 0.177 & eng\_Latn & 0.451 & cmn\_Hani & 0.275 & \multirow{6}{*}{\rotatebox[origin=c]{90}{POS}} & jpn\_Jpan & 0.165 & eng\_Latn & 0.534 & cmn\_Hani & 0.369 \\
& & kir\_Cyrl & 0.391 & eng\_Latn & 0.564 & rus\_Cyrl & 0.173 & & mlt\_Latn & 0.603 & arb\_Arab & 0.798 & spa\_Latn & 0.196 \\
& & mya\_Mymr & 0.455 & cmn\_Hani & 0.607 & hin\_Deva & 0.153 & & wol\_Latn & 0.606 & eng\_Latn & 0.679 & spa\_Latn & 0.074 \\
    \cmidrule{1-1} \cmidrule{3-8} \cmidrule{10-15}
    \multirow{3}{*}{\rotatebox[origin=c]{90}{low end}} & & pes\_Arab & 0.653 & hin\_Deva & 0.606 & arb\_Arab & -0.047 & & ekk\_Latn & 0.815 & eng\_Latn & 0.790 & rus\_Cyrl & -0.025 \\
& & tgl\_Latn & 0.745 & eng\_Latn & 0.667 & spa\_Latn & -0.078 & & bam\_Latn & 0.451 & eng\_Latn & 0.411 & spa\_Latn & -0.039 \\
& & sun\_Latn & 0.577 & eng\_Latn & 0.490 & spa\_Latn & -0.087 & & gla\_Latn & 0.588 & rus\_Cyrl & 0.548 & spa\_Latn & -0.040 \\
    \midrule
    \midrule    
    \multirow{3}{*}{\rotatebox[origin=c]{90}{high end}} & \multirow{6}{*}{\rotatebox[origin=c]{90}{MASSIVE}} & mya\_Mymr & 0.616 & cmn\_Hani & 0.707 & hin\_Deva & 0.091 & \multirow{6}{*}{\rotatebox[origin=c]{90}{Taxi1500}} & tgk\_Cyrl & 0.493 & hin\_Deva & 0.724 & rus\_Cyrl & 0.231 \\
& & amh\_Ethi & 0.532 & arb\_Arab & 0.611 & hin\_Deva & 0.079 & & kin\_Latn & 0.431 & eng\_Latn & 0.619 & spa\_Latn & 0.188 \\
& & jpn\_Jpan & 0.384 & eng\_Latn & 0.448 & cmn\_Hani & 0.064 & & kik\_Latn & 0.384 & eng\_Latn & 0.555 & spa\_Latn & 0.172 \\
    \cmidrule{1-1} \cmidrule{3-8} \cmidrule{10-15}
    \multirow{3}{*}{\rotatebox[origin=c]{90}{low end}} & & cym\_Latn & 0.495 & rus\_Cyrl & 0.480 & spa\_Latn & -0.015 & & ckb\_Arab & 0.622 & hin\_Deva & 0.539 & arb\_Arab & -0.083 \\
& & tgl\_Latn & 0.752 & eng\_Latn & 0.723 & spa\_Latn & -0.028 & & nld\_Latn & 0.713 & eng\_Latn & 0.628 & spa\_Latn & -0.085 \\
& & deu\_Latn & 0.759 & eng\_Latn & 0.726 & spa\_Latn & -0.033 & & kac\_Latn & 0.580 & cmn\_Hani & 0.483 & hin\_Deva & -0.097 \\
    \bottomrule
        \end{tabular}
    }
    \caption{Results for three languages each with the largest (high end) and smallest (low end) gains from \mPLMMeasure vs. GEN for four tasks.
    \mPLMMeasure's gain over GEN is large at the high end and smaller negative at the low end.
    We report both the selected source languages and the results on the evaluated target languages. For \mPLMMeasure, the results are derived from the layers exhibiting the best performances as shown in Fig.~\ref{fig:cross-lingual result}. See \S\ref{sec:results} for detailed results for each task and each target language.}
    \label{tab:crossling}
\end{table*}

The macro average results of cross-lingual transfer across target languages for both \mPLMMeasure and baselines are presented in Fig.~\ref{fig:cross-lingual result}. Among the evaluated tasks, ENG exhibits the worst
performance in three out of four tasks, emphasizing the
importance of considering language similarity when selecting
source languages for cross-lingual transfer. \mPLMMeasure
surpasses all linguistic similarity measures in every task,
including both syntactic and semantic tasks, across all
layers except layer 0. This indicates that \mPLMMeasure is
more effective in selecting source languages that enhance
the performance of target languages compared to linguistic
similarity measures.

For low-level syntactic
tasks, the lower layers (layer 1 or 2) exhibit superior
performance compared to all other layers. Conversely, for
high-level semantic tasks, it is the middle layer of the \mPLM that consistently achieves the highest results across all layers. This can be attributed to its ability to capture intricate similarity patterns.

In Tab.~\ref{tab:crossling},
we further explore the benefits of \mPLMMeasure in
cross-lingual transfer.  We present a comprehensive analysis of the top 3
performance improvements and declines across languages. We compare  \mPLMMeasure and GEN across four cross-lingual transfer tasks. By examining these results, we gain deeper insights into the advantages of \mPLMMeasure in facilitating effective cross-lingual transfer.

The results clearly demonstrate that \mPLMMeasure has a substantial performance advantage over GEN for certain target languages. On one hand, for languages without any source language in the same language family, such as Japanese (jpn\_Jpan), \mPLMMeasure successfully identifies its similar language, Chinese (cmn\_Hani), whereas GEN fails to do so. Notably, in the case of Japanese, \mPLMMeasure outperforms GEN by 27.5\% for NER, 36.9\% for POS, and 6.4\% for MASSIVE.

On the other hand, for languages having source languages within the same language family, \mPLMMeasure accurately detects the appropriate source language, leading to improved cross-lingual transfer performance. In the case of Burmese (mya\_Mymr), \mPLMMeasure accurately identifies Hindi (hin\_Deva) as the source language, while GEN mistakenly selects Chinese (cmn\_Hani). This distinction results in a significant performance improvement of 15.3\% for NER and 9.1\% for MASSIVE.

However, we also observe that \mPLMMeasure falls short for certain languages when compared to GEN, although the losses are smaller in magnitude compared to the improvements. This finding suggests that achieving better performance in cross-lingual transfer is not solely dependent on language similarity. As mentioned in previous studies such as \citet{DBLP:conf/emnlp/LauscherRVG20} and \citet{DBLP:journals/corr/abs-2212-09651}, the size of the pretraining data for the source languages also plays a crucial role in cross-lingual transfer.

\section{Related Work}

\subsection{Language Typology and Clustering}

Similarity between languages can be
due to common ancestry
in the genealogical language tree, but also influenced by
linguistic influence and
borrowing \citep{ad01,haspelmath_martin_2004_580172}. Linguists
have conducted extensive relevant research by constructing high-quality
typological, geographical, and phylogenetic databases,
including WALS \citep{dryer2013world},
Glottolog \citep{hammarstrom2017glottolog},
Ethnologue \citep{saggion2023findings}, and
PHOIBLE \citep{moran2014phoible,phoible}. The lang2vec
tool \citep{DBLP:conf/eacl/LevinLMLKT17} further integrates
these datasets into multiple linguistic distances. Despite
its integration of multiple linguistic measures, lang2vec
weights each measure equally, and the quantification of
these measures for language similarity computation remains a
challenge.

In addition to linguistic measures, some non-lingustic measures are also proposed to measure similarity between languages. Specifically, \citet{holman2011automated} use Levenshtein (edit) distance to compute the lexical similarity between languages. \citet{DBLP:conf/acl/LinCLLZXRHZMALN19} propose dataset-dependent features, which are statistical features specific to the corpus used, e.g., lexical overlap. \citet{DBLP:journals/corr/abs-2305-13401} measure language similarity with basic concepts across languages. However, these methods fail to capture deeper similarities beyond surface-level features.

Language representation is another important category of
language similarity
measures.
Before the era of multilingual pretrained language models (mPLMs), exploiting distributed language representations for measuring language similarity have been studied \citep{DBLP:conf/eacl/TiedemannO17,DBLP:conf/naacl/BjervaA18}. Recent mPLMs trained with massive data have become a new standard for multilingual representation learning.
\citet{DBLP:conf/emnlp/TanCHXQL19} represent each
language by an embedding vector and cluster them in the
embedding space. \citet{DBLP:conf/emnlp/FanLMHL0D21} find
the representation sprachbund of \mPLMs, and then train
separate \mPLMs for each sprachbund. However, these studies do not delve into the research questions mentioned in \S\ref{sec:intro}, and it motivates us to carry out a comprehensive investigation of language similarity using \mPLMs.

\subsection{Multilingual Pretrained Language Models}

The advent of \mPLMs, e.g., mBERT \citep{DBLP:conf/naacl/DevlinCLT19}, XLM \citep{DBLP:conf/nips/ConneauL19}, and XLM-R \citep{DBLP:conf/acl/ConneauKGCWGGOZ20}, have brought significant performance gains on numerous multilingual natural language understanding benchmarks \citep{DBLP:conf/icml/HuRSNFJ20}.

Given their success, a variety of following \mPLMs are proposed. Specifically, different architectures, including decoder-only, e.g., mGPT \citep{DBLP:journals/corr/abs-2204-07580} and BLOOM \citep{DBLP:journals/corr/abs-2211-05100}, and encoder-decoder, e.g., mT5 \citep{DBLP:conf/naacl/XueCRKASBR21}, are designed. Tokenizer-free models, including CANINE \citep{DBLP:journals/tacl/ClarkGTW22}, ByT5 \citep{DBLP:journals/tacl/XueBCANKRR22}, and Charformer \citep{DBLP:conf/iclr/Tay0RGCB0B0M22}, are also proposed. \citet{DBLP:journals/tacl/ClarkGTW22} introduce  CANINE-S and CANINE-C. CANINE-S adopts a subword-based loss, while CANINE-C uses a character-based one. Glot500 \citep{imanigooghari2023glot500} extends XLM-R to cover more than 500 languages using vocabulary extension and continued pretraining. Both InfoXLM \citep{DBLP:conf/naacl/ChiDWYSWSMHZ21} and XLM-Align \citep{DBLP:conf/acl/Chi0ZHMHW20} exploit parallel objectives to further improve \mPLMs.
Some \mPLMs are specifically proposed for Machine Translation, e.g., M2M-100 \citep{DBLP:journals/jmlr/FanBSMEGBCWCGBL21} and NLLB-200 \citep{DBLP:journals/corr/abs-2207-04672}.
XLS-R-300M \citep{DBLP:journals/corr/abs-2111-09296}
is a speech (as opposed to text) model.

Follow-up works show that strong language-specific signals
are encoded in \mPLMs by means of probing tasks \citep{DBLP:conf/emnlp/WuD19,DBLP:conf/coling/RamaBE20,DBLP:conf/acl/PiresSG19,DBLP:conf/eacl/MullerESS21,DBLP:journals/corr/abs-2109-08040,DBLP:journals/coling/ChoenniS22} and investigating the geometry of \mPLMs \citep{DBLP:conf/emnlp/LibovickyRF20,DBLP:journals/corr/abs-2205-10964,DBLP:journals/corr/abs-2305-00090}. Concurrent with our work, \citet{DBLP:journals/corr/abs-2305-02151} have verified that the language representations encoded in mBERT correlate with both linguistic typology and cross-lingual transfer on XNLI for 15 languages. However, these methods lack in-depth analysis and investigate on a limited set of mPLMs and downstream tasks. This inspires us to conduct quantitative and qualitative analysis on linguistic typology and cross-lingual transfer with a broad and diverse set of \mPLMs and downstream tasks.

\section{Conclusion}

In this paper, we introduce \mPLMMeasure, a novel approach for measuring language similarities. Extensive experiments substantiate the superior performance of \mPLMMeasure compared to linguistic similarity measures. Our study reveals variations in similarity results across different \mPLMs and layers within an \mPLM. Furthermore, our findings reveal that \mPLMMeasure effectively identifies the source language to enhance cross-lingual transfer.

The results obtained from \mPLMMeasure have significant
implications for multilinguality. On the one hand, it can be
further used in linguistic study and downstream
applications, such as cross-lingual transfer, as elaborated in the paper. On the other hand, these findings provide valuable insights for improving \mPLMs, offering opportunities for their further development and enhancement.

\section*{Limitations}

(1) The performance of \mPLMMeasure may be strongly influenced by the quality and quantity of data used for training \mPLMs, as well as the degree to which the target language can be accurately represented.
(2) The success of mPLM-Sim depends on the supporting languages of mPLMs. We conduct further experiment and analysis at \S\ref{sec:unseen}.
(3) As for \S\ref{sec:cmp_model}, we are unable to conduct a strictly fair comparison due to the varying settings in which \mPLMs are pretrained, including the use of different corpora and model sizes.

\section*{Acknowledgements}
This work was funded by the European Research Council (NonSequeToR, grant \#740516, and DECOLLAGE, ERC-2022-CoG \#101088763),  EU's Horizon Europe Research and Innovation Actions (UTTER, contract 101070631), by Fundação para a Ciência e Tecnologia through contract UIDB/50008/2020,  by the DAAD programme Konrad Zuse Schools of Excellence in Artificial Intelligence, sponsored by the Federal Ministry of Education and Research, and by the Portuguese Recovery and Resilience Plan through project C645008882-00000055 (Center for Responsible AI). Peiqin Lin acknowledges travel support from ELISE (GA no 951847).

\bibliography{anthology,custom}

\begin{thebibliography}{55}
\expandafter\ifx\csname natexlab\endcsname\relax\def\natexlab#1{#1}\fi

\bibitem[{Abdullah et~al.(2023)Abdullah, Shaik, and
  Klakow}]{abdullah2023nature}
Badr~M Abdullah, Mohammed~Maqsood Shaik, and Dietrich Klakow. 2023.
\newblock On the nature of discrete speech representations in multilingual
  self-supervised models.
\newblock In \emph{Proceedings of the 5th Workshop on Research in Computational
  Linguistic Typology and Multilingual NLP}, pages 159--161.

\bibitem[{Aikhenvald and Dixon(2001)}]{ad01}
Alexandra~Y. Aikhenvald and R.~M.~W. Dixon. 2001.
\newblock \emph{Areal diffusion and genetic inheritance}.
\newblock Oxford University Press, Oxford.

\bibitem[{Babu et~al.(2021)Babu, Wang, Tjandra, Lakhotia, Xu, Goyal, Singh, von
  Platen, Saraf, Pino, Baevski, Conneau, and
  Auli}]{DBLP:journals/corr/abs-2111-09296}
Arun Babu, Changhan Wang, Andros Tjandra, Kushal Lakhotia, Qiantong Xu, Naman
  Goyal, Kritika Singh, Patrick von Platen, Yatharth Saraf, Juan Pino, Alexei
  Baevski, Alexis Conneau, and Michael Auli. 2021.
\newblock \href {http://arxiv.org/abs/2111.09296} {{XLS-R:} self-supervised
  cross-lingual speech representation learning at scale}.
\newblock \emph{CoRR}, abs/2111.09296.

\bibitem[{Bjerva and Augenstein(2018)}]{DBLP:conf/naacl/BjervaA18}
Johannes Bjerva and Isabelle Augenstein. 2018.
\newblock \href {https://doi.org/10.18653/V1/N18-1083} {From phonology to
  syntax: Unsupervised linguistic typology at different levels with language
  embeddings}.
\newblock In \emph{Proceedings of the 2018 Conference of the North American
  Chapter of the Association for Computational Linguistics: Human Language
  Technologies, {NAACL-HLT} 2018, New Orleans, Louisiana, USA, June 1-6, 2018,
  Volume 1 (Long Papers)}, pages 907--916. Association for Computational
  Linguistics.

\bibitem[{Chang et~al.(2022)Chang, Tu, and
  Bergen}]{DBLP:journals/corr/abs-2205-10964}
Tyler~A. Chang, Zhuowen Tu, and Benjamin~K. Bergen. 2022.
\newblock \href {https://doi.org/10.48550/arXiv.2205.10964} {The geometry of
  multilingual language model representations}.
\newblock \emph{CoRR}, abs/2205.10964.

\bibitem[{Chi et~al.(2021{\natexlab{a}})Chi, Dong, Wei, Yang, Singhal, Wang,
  Song, Mao, Huang, and Zhou}]{DBLP:conf/naacl/ChiDWYSWSMHZ21}
Zewen Chi, Li~Dong, Furu Wei, Nan Yang, Saksham Singhal, Wenhui Wang, Xia Song,
  Xian{-}Ling Mao, Heyan Huang, and Ming Zhou. 2021{\natexlab{a}}.
\newblock \href {https://doi.org/10.18653/v1/2021.naacl-main.280} {Infoxlm: An
  information-theoretic framework for cross-lingual language model
  pre-training}.
\newblock In \emph{Proceedings of the 2021 Conference of the North American
  Chapter of the Association for Computational Linguistics: Human Language
  Technologies, {NAACL-HLT} 2021, Online, June 6-11, 2021}, pages 3576--3588.
  Association for Computational Linguistics.

\bibitem[{Chi et~al.(2021{\natexlab{b}})Chi, Dong, Zheng, Huang, Mao, Huang,
  and Wei}]{DBLP:conf/acl/Chi0ZHMHW20}
Zewen Chi, Li~Dong, Bo~Zheng, Shaohan Huang, Xian{-}Ling Mao, Heyan Huang, and
  Furu Wei. 2021{\natexlab{b}}.
\newblock \href {https://doi.org/10.18653/v1/2021.acl-long.265} {Improving
  pretrained cross-lingual language models via self-labeled word alignment}.
\newblock In \emph{Proceedings of the 59th Annual Meeting of the Association
  for Computational Linguistics and the 11th International Joint Conference on
  Natural Language Processing, {ACL/IJCNLP} 2021, (Volume 1: Long Papers),
  Virtual Event, August 1-6, 2021}, pages 3418--3430. Association for
  Computational Linguistics.

\bibitem[{Choenni and Shutova(2022)}]{DBLP:journals/coling/ChoenniS22}
Rochelle Choenni and Ekaterina Shutova. 2022.
\newblock \href {https://doi.org/10.1162/coli\_a\_00444} {Investigating
  language relationships in multilingual sentence encoders through the lens of
  linguistic typology}.
\newblock \emph{Comput. Linguistics}, 48(3):635--672.

\bibitem[{Clark et~al.(2022)Clark, Garrette, Turc, and
  Wieting}]{DBLP:journals/tacl/ClarkGTW22}
Jonathan~H. Clark, Dan Garrette, Iulia Turc, and John Wieting. 2022.
\newblock \href {https://doi.org/10.1162/tacl\_a\_00448} {Canine: Pre-training
  an efficient tokenization-free encoder for language representation}.
\newblock \emph{Trans. Assoc. Comput. Linguistics}, 10:73--91.

\bibitem[{Conneau et~al.(2020)Conneau, Khandelwal, Goyal, Chaudhary, Wenzek,
  Guzm{\'{a}}n, Grave, Ott, Zettlemoyer, and
  Stoyanov}]{DBLP:conf/acl/ConneauKGCWGGOZ20}
Alexis Conneau, Kartikay Khandelwal, Naman Goyal, Vishrav Chaudhary, Guillaume
  Wenzek, Francisco Guzm{\'{a}}n, Edouard Grave, Myle Ott, Luke Zettlemoyer,
  and Veselin Stoyanov. 2020.
\newblock \href {https://doi.org/10.18653/v1/2020.acl-main.747} {Unsupervised
  cross-lingual representation learning at scale}.
\newblock In \emph{Proceedings of the 58th Annual Meeting of the Association
  for Computational Linguistics, {ACL} 2020, Online, July 5-10, 2020}, pages
  8440--8451. Association for Computational Linguistics.

\bibitem[{Conneau and Lample(2019)}]{DBLP:conf/nips/ConneauL19}
Alexis Conneau and Guillaume Lample. 2019.
\newblock \href
  {https://proceedings.neurips.cc/paper/2019/hash/c04c19c2c2474dbf5f7ac4372c5b9af1-Abstract.html}
  {Cross-lingual language model pretraining}.
\newblock In \emph{Advances in Neural Information Processing Systems 32: Annual
  Conference on Neural Information Processing Systems 2019, NeurIPS 2019,
  December 8-14, 2019, Vancouver, BC, Canada}, pages 7057--7067.

\bibitem[{Conneau et~al.(2022)Conneau, Ma, Khanuja, Zhang, Axelrod, Dalmia,
  Riesa, Rivera, and Bapna}]{DBLP:journals/corr/abs-2205-12446}
Alexis Conneau, Min Ma, Simran Khanuja, Yu~Zhang, Vera Axelrod, Siddharth
  Dalmia, Jason Riesa, Clara Rivera, and Ankur Bapna. 2022.
\newblock \href {https://doi.org/10.48550/arXiv.2205.12446} {{FLEURS:} few-shot
  learning evaluation of universal representations of speech}.
\newblock \emph{CoRR}, abs/2205.12446.

\bibitem[{Costa{-}juss{\`{a}} et~al.(2022)Costa{-}juss{\`{a}}, Cross,
  {\c{C}}elebi, Elbayad, Heafield, Heffernan, Kalbassi, Lam, Licht, Maillard,
  Sun, Wang, Wenzek, Youngblood, Akula, Barrault, Gonzalez, Hansanti, Hoffman,
  Jarrett, Sadagopan, Rowe, Spruit, Tran, Andrews, Ayan, Bhosale, Edunov, Fan,
  Gao, Goswami, Guzm{\'{a}}n, Koehn, Mourachko, Ropers, Saleem, Schwenk, and
  Wang}]{DBLP:journals/corr/abs-2207-04672}
Marta~R. Costa{-}juss{\`{a}}, James Cross, Onur {\c{C}}elebi, Maha Elbayad,
  Kenneth Heafield, Kevin Heffernan, Elahe Kalbassi, Janice Lam, Daniel Licht,
  Jean Maillard, Anna Sun, Skyler Wang, Guillaume Wenzek, Al~Youngblood, Bapi
  Akula, Lo{\"{\i}}c Barrault, Gabriel~Mejia Gonzalez, Prangthip Hansanti, John
  Hoffman, Semarley Jarrett, Kaushik~Ram Sadagopan, Dirk Rowe, Shannon Spruit,
  Chau Tran, Pierre Andrews, Necip~Fazil Ayan, Shruti Bhosale, Sergey Edunov,
  Angela Fan, Cynthia Gao, Vedanuj Goswami, Francisco Guzm{\'{a}}n, Philipp
  Koehn, Alexandre Mourachko, Christophe Ropers, Safiyyah Saleem, Holger
  Schwenk, and Jeff Wang. 2022.
\newblock \href {https://doi.org/10.48550/arXiv.2207.04672} {No language left
  behind: Scaling human-centered machine translation}.
\newblock \emph{CoRR}, abs/2207.04672.

\bibitem[{de~Marneffe et~al.(2021)de~Marneffe, Manning, Nivre, and
  Zeman}]{DBLP:journals/coling/MarneffeMNZ21}
Marie{-}Catherine de~Marneffe, Christopher~D. Manning, Joakim Nivre, and Daniel
  Zeman. 2021.
\newblock \href {https://doi.org/10.1162/coli\_a\_00402} {Universal
  dependencies}.
\newblock \emph{Comput. Linguistics}, 47(2):255--308.

\bibitem[{Devlin et~al.(2019)Devlin, Chang, Lee, and
  Toutanova}]{DBLP:conf/naacl/DevlinCLT19}
Jacob Devlin, Ming{-}Wei Chang, Kenton Lee, and Kristina Toutanova. 2019.
\newblock \href {https://doi.org/10.18653/v1/n19-1423} {{BERT:} pre-training of
  deep bidirectional transformers for language understanding}.
\newblock In \emph{Proceedings of the 2019 Conference of the North American
  Chapter of the Association for Computational Linguistics: Human Language
  Technologies, {NAACL-HLT} 2019, Minneapolis, MN, USA, June 2-7, 2019, Volume
  1 (Long and Short Papers)}, pages 4171--4186. Association for Computational
  Linguistics.

\bibitem[{Dryer and Haspelmath(2013)}]{dryer2013world}
Matthew~S Dryer and Martin Haspelmath. 2013.
\newblock The world atlas of language structures online.

\bibitem[{Fan et~al.(2021{\natexlab{a}})Fan, Bhosale, Schwenk, Ma, El{-}Kishky,
  Goyal, Baines, Celebi, Wenzek, Chaudhary, Goyal, Birch, Liptchinsky, Edunov,
  Auli, and Joulin}]{DBLP:journals/jmlr/FanBSMEGBCWCGBL21}
Angela Fan, Shruti Bhosale, Holger Schwenk, Zhiyi Ma, Ahmed El{-}Kishky,
  Siddharth Goyal, Mandeep Baines, Onur Celebi, Guillaume Wenzek, Vishrav
  Chaudhary, Naman Goyal, Tom Birch, Vitaliy Liptchinsky, Sergey Edunov,
  Michael Auli, and Armand Joulin. 2021{\natexlab{a}}.
\newblock \href {http://jmlr.org/papers/v22/20-1307.html} {Beyond
  english-centric multilingual machine translation}.
\newblock \emph{J. Mach. Learn. Res.}, 22:107:1--107:48.

\bibitem[{Fan et~al.(2021{\natexlab{b}})Fan, Liang, Muzio, Hassan, Li, Zhou,
  and Duan}]{DBLP:conf/emnlp/FanLMHL0D21}
Yimin Fan, Yaobo Liang, Alexandre Muzio, Hany Hassan, Houqiang Li, Ming Zhou,
  and Nan Duan. 2021{\natexlab{b}}.
\newblock \href {https://doi.org/10.18653/v1/2021.findings-emnlp.75}
  {Discovering representation sprachbund for multilingual pre-training}.
\newblock In \emph{Findings of the Association for Computational Linguistics:
  {EMNLP} 2021, Virtual Event / Punta Cana, Dominican Republic, 16-20 November,
  2021}, pages 881--894. Association for Computational Linguistics.

\bibitem[{FitzGerald et~al.(2022)FitzGerald, Hench, Peris, Mackie, Rottmann,
  Sanchez, Nash, Urbach, Kakarala, Singh, Ranganath, Crist, Britan, Leeuwis,
  T{\"{u}}r, and Natarajan}]{DBLP:journals/corr/abs-2204-08582}
Jack FitzGerald, Christopher Hench, Charith Peris, Scott Mackie, Kay Rottmann,
  Ana Sanchez, Aaron Nash, Liam Urbach, Vishesh Kakarala, Richa Singh, Swetha
  Ranganath, Laurie Crist, Misha Britan, Wouter Leeuwis, G{\"{o}}khan
  T{\"{u}}r, and Prem Natarajan. 2022.
\newblock \href {https://doi.org/10.48550/arXiv.2204.08582} {{MASSIVE:} {A}
  1m-example multilingual natural language understanding dataset with 51
  typologically-diverse languages}.
\newblock \emph{CoRR}, abs/2204.08582.

\bibitem[{Hammarstr{\"o}m et~al.(2017)Hammarstr{\"o}m, Forkel, and
  Haspelmath}]{hammarstrom2017glottolog}
Harald Hammarstr{\"o}m, Robert Forkel, and Martin Haspelmath. 2017.
\newblock Glottolog 3.0.
\newblock \emph{Max Planck Institute for the Science of Human History}.

\bibitem[{Haspelmath(2004)}]{haspelmath_martin_2004_580172}
Martin Haspelmath. 2004.
\newblock \href {https://doi.org/10.1075/sl.28.1.10has} {{How hopeless is
  genealogical linguistics, and how advanced is areal linguistics?}}
\newblock \emph{Studies in Language}, 28(1):209--223.

\bibitem[{Holman et~al.(2011)Holman, Brown, Wichmann, M{\"u}ller, Velupillai,
  Hammarstr{\"o}m, Sauppe, Jung, Bakker, Brown et~al.}]{holman2011automated}
Eric~W Holman, Cecil~H Brown, S{\o}ren Wichmann, Andr{\'e} M{\"u}ller, Viveka
  Velupillai, Harald Hammarstr{\"o}m, Sebastian Sauppe, Hagen Jung, Dik Bakker,
  Pamela Brown, et~al. 2011.
\newblock Automated dating of the world’s language families based on lexical
  similarity.
\newblock \emph{Current Anthropology}, 52(6):841--875.

\bibitem[{Hu et~al.(2020)Hu, Ruder, Siddhant, Neubig, Firat, and
  Johnson}]{DBLP:conf/icml/HuRSNFJ20}
Junjie Hu, Sebastian Ruder, Aditya Siddhant, Graham Neubig, Orhan Firat, and
  Melvin Johnson. 2020.
\newblock \href {http://proceedings.mlr.press/v119/hu20b.html} {{XTREME:} {A}
  massively multilingual multi-task benchmark for evaluating cross-lingual
  generalisation}.
\newblock In \emph{Proceedings of the 37th International Conference on Machine
  Learning, {ICML} 2020, 13-18 July 2020, Virtual Event}, volume 119 of
  \emph{Proceedings of Machine Learning Research}, pages 4411--4421. {PMLR}.

\bibitem[{Imai et~al.(2023)Imai, Kawahara, Orita, and
  Oda}]{DBLP:conf/acl/ImaiKOO23}
Sakura Imai, Daisuke Kawahara, Naho Orita, and Hiromune Oda. 2023.
\newblock \href {https://doi.org/10.18653/V1/2023.ACL-SRW.24} {Theoretical
  linguistics rivals embeddings in language clustering for multilingual named
  entity recognition}.
\newblock In \emph{Proceedings of the 61st Annual Meeting of the Association
  for Computational Linguistics: Student Research Workshop, {ACL} 2023,
  Toronto, Canada, July 9-14, 2023}, pages 139--151. Association for
  Computational Linguistics.

\bibitem[{Imani et~al.(2023)Imani, Lin, Kargaran, Severini, Sabet, Kassner, Ma,
  Schmid, Martins, Yvon, and Schütze}]{imanigooghari2023glot500}
Ayyoob Imani, Peiqin Lin, Amir~Hossein Kargaran, Silvia Severini, Masoud~Jalili
  Sabet, Nora Kassner, Chunlan Ma, Helmut Schmid, André F.~T. Martins,
  François Yvon, and Hinrich Schütze. 2023.
\newblock \href {http://arxiv.org/abs/2305.12182} {Glot500: Scaling
  multilingual corpora and language models to 500 languages}.

\bibitem[{Jawahar et~al.(2019)Jawahar, Sagot, and
  Seddah}]{DBLP:conf/acl/JawaharSS19}
Ganesh Jawahar, Beno{\^{\i}}t Sagot, and Djam{\'{e}} Seddah. 2019.
\newblock \href {https://doi.org/10.18653/v1/p19-1356} {What does {BERT} learn
  about the structure of language?}
\newblock In \emph{Proceedings of the 57th Conference of the Association for
  Computational Linguistics, {ACL} 2019, Florence, Italy, July 28- August 2,
  2019, Volume 1: Long Papers}, pages 3651--3657. Association for Computational
  Linguistics.

\bibitem[{Lauscher et~al.(2020)Lauscher, Ravishankar, Vulic, and
  Glavas}]{DBLP:conf/emnlp/LauscherRVG20}
Anne Lauscher, Vinit Ravishankar, Ivan Vulic, and Goran Glavas. 2020.
\newblock \href {https://doi.org/10.18653/v1/2020.emnlp-main.363} {From zero to
  hero: On the limitations of zero-shot language transfer with multilingual
  transformers}.
\newblock In \emph{Proceedings of the 2020 Conference on Empirical Methods in
  Natural Language Processing, {EMNLP} 2020, Online, November 16-20, 2020},
  pages 4483--4499. Association for Computational Linguistics.

\bibitem[{Liang et~al.(2021)Liang, Dufter, and
  Sch{\"{u}}tze}]{DBLP:journals/corr/abs-2109-08040}
Sheng Liang, Philipp Dufter, and Hinrich Sch{\"{u}}tze. 2021.
\newblock \href {http://arxiv.org/abs/2109.08040} {Locating language-specific
  information in contextualized embeddings}.
\newblock \emph{CoRR}, abs/2109.08040.

\bibitem[{Libovick{\'{y}} et~al.(2020)Libovick{\'{y}}, Rosa, and
  Fraser}]{DBLP:conf/emnlp/LibovickyRF20}
Jindrich Libovick{\'{y}}, Rudolf Rosa, and Alexander Fraser. 2020.
\newblock \href {https://doi.org/10.18653/v1/2020.findings-emnlp.150} {On the
  language neutrality of pre-trained multilingual representations}.
\newblock In \emph{Findings of the Association for Computational Linguistics:
  {EMNLP} 2020, Online Event, 16-20 November 2020}, volume {EMNLP} 2020 of
  \emph{Findings of {ACL}}, pages 1663--1674. Association for Computational
  Linguistics.

\bibitem[{Lin et~al.(2019)Lin, Chen, Lee, Li, Zhang, Xia, Rijhwani, He, Zhang,
  Ma, Anastasopoulos, Littell, and Neubig}]{DBLP:conf/acl/LinCLLZXRHZMALN19}
Yu{-}Hsiang Lin, Chian{-}Yu Chen, Jean Lee, Zirui Li, Yuyan Zhang, Mengzhou
  Xia, Shruti Rijhwani, Junxian He, Zhisong Zhang, Xuezhe Ma, Antonios
  Anastasopoulos, Patrick Littell, and Graham Neubig. 2019.
\newblock \href {https://doi.org/10.18653/v1/p19-1301} {Choosing transfer
  languages for cross-lingual learning}.
\newblock In \emph{Proceedings of the 57th Conference of the Association for
  Computational Linguistics, {ACL} 2019, Florence, Italy, July 28- August 2,
  2019, Volume 1: Long Papers}, pages 3125--3135. Association for Computational
  Linguistics.

\bibitem[{Littell et~al.(2017)Littell, Mortensen, Lin, Kairis, Turner, and
  Levin}]{DBLP:conf/eacl/LevinLMLKT17}
Patrick Littell, David~R. Mortensen, Ke~Lin, Katherine Kairis, Carlisle Turner,
  and Lori~S. Levin. 2017.
\newblock \href {https://doi.org/10.18653/v1/e17-2002} {{URIEL} and lang2vec:
  Representing languages as typological, geographical, and phylogenetic
  vectors}.
\newblock In \emph{Proceedings of the 15th Conference of the European Chapter
  of the Association for Computational Linguistics, {EACL} 2017, Valencia,
  Spain, April 3-7, 2017, Volume 2: Short Papers}, pages 8--14. Association for
  Computational Linguistics.

\bibitem[{Ma et~al.(2023)Ma, ImaniGooghari, Ye, Asgari, and
  Schütze}]{ma2023taxi1500}
Chunlan Ma, Ayyoob ImaniGooghari, Haotian Ye, Ehsaneddin Asgari, and Hinrich
  Schütze. 2023.
\newblock \href {http://arxiv.org/abs/2305.08487} {Taxi1500: A multilingual
  dataset for text classification in 1500 languages}.

\bibitem[{Mayer and Cysouw(2014)}]{DBLP:conf/lrec/MayerC14}
Thomas Mayer and Michael Cysouw. 2014.
\newblock \href
  {http://www.lrec-conf.org/proceedings/lrec2014/summaries/220.html} {Creating
  a massively parallel bible corpus}.
\newblock In \emph{Proceedings of the Ninth International Conference on
  Language Resources and Evaluation, {LREC} 2014, Reykjavik, Iceland, May
  26-31, 2014}, pages 3158--3163. European Language Resources Association
  {(ELRA)}.

\bibitem[{Moran and McCloy(2019)}]{phoible}
Steven Moran and Daniel McCloy, editors. 2019.
\newblock \href {https://phoible.org/} {\emph{PHOIBLE 2.0}}.
\newblock Max Planck Institute for the Science of Human History, Jena.

\bibitem[{Moran et~al.(2014)Moran, McCloy, and Wright}]{moran2014phoible}
Steven Moran, Daniel McCloy, and Richard Wright. 2014.
\newblock Phoible online.

\bibitem[{Muennighoff(2022)}]{DBLP:journals/corr/abs-2202-08904}
Niklas Muennighoff. 2022.
\newblock \href {http://arxiv.org/abs/2202.08904} {{SGPT:} {GPT} sentence
  embeddings for semantic search}.
\newblock \emph{CoRR}, abs/2202.08904.

\bibitem[{M{\"{u}}ller et~al.(2021)M{\"{u}}ller, Elazar, Sagot, and
  Seddah}]{DBLP:conf/eacl/MullerESS21}
Benjamin M{\"{u}}ller, Yanai Elazar, Beno{\^{\i}}t Sagot, and Djam{\'{e}}
  Seddah. 2021.
\newblock \href {https://doi.org/10.18653/v1/2021.eacl-main.189} {First align,
  then predict: Understanding the cross-lingual ability of multilingual
  {BERT}}.
\newblock In \emph{Proceedings of the 16th Conference of the European Chapter
  of the Association for Computational Linguistics: Main Volume, {EACL} 2021,
  Online, April 19 - 23, 2021}, pages 2214--2231. Association for Computational
  Linguistics.

\bibitem[{Nie et~al.(2022)Nie, Liang, Schmid, and
  Sch{\"{u}}tze}]{DBLP:journals/corr/abs-2212-09651}
Ercong Nie, Sheng Liang, Helmut Schmid, and Hinrich Sch{\"{u}}tze. 2022.
\newblock \href {https://doi.org/10.48550/arXiv.2212.09651} {Cross-lingual
  retrieval augmented prompt for low-resource languages}.
\newblock \emph{CoRR}, abs/2212.09651.

\bibitem[{{\"{O}}stling and Tiedemann(2017)}]{DBLP:conf/eacl/TiedemannO17}
Robert {\"{O}}stling and J{\"{o}}rg Tiedemann. 2017.
\newblock \href {https://doi.org/10.18653/V1/E17-2102} {Continuous
  multilinguality with language vectors}.
\newblock In \emph{Proceedings of the 15th Conference of the European Chapter
  of the Association for Computational Linguistics, {EACL} 2017, Valencia,
  Spain, April 3-7, 2017, Volume 2: Short Papers}, pages 644--649. Association
  for Computational Linguistics.

\bibitem[{Pan et~al.(2017)Pan, Zhang, May, Nothman, Knight, and
  Ji}]{DBLP:conf/acl/PanZMNKJ17}
Xiaoman Pan, Boliang Zhang, Jonathan May, Joel Nothman, Kevin Knight, and Heng
  Ji. 2017.
\newblock \href {https://doi.org/10.18653/v1/P17-1178} {Cross-lingual name
  tagging and linking for 282 languages}.
\newblock In \emph{Proceedings of the 55th Annual Meeting of the Association
  for Computational Linguistics, {ACL} 2017, Vancouver, Canada, July 30 -
  August 4, Volume 1: Long Papers}, pages 1946--1958. Association for
  Computational Linguistics.

\bibitem[{Philippy et~al.(2023)Philippy, Guo, and
  Haddadan}]{DBLP:journals/corr/abs-2305-02151}
Fred Philippy, Siwen Guo, and Shohreh Haddadan. 2023.
\newblock \href {https://doi.org/10.48550/ARXIV.2305.02151} {Identifying the
  correlation between language distance and cross-lingual transfer in a
  multilingual representation space}.
\newblock \emph{CoRR}, abs/2305.02151.

\bibitem[{Pires et~al.(2019)Pires, Schlinger, and
  Garrette}]{DBLP:conf/acl/PiresSG19}
Telmo Pires, Eva Schlinger, and Dan Garrette. 2019.
\newblock \href {https://doi.org/10.18653/v1/p19-1493} {How multilingual is
  multilingual bert?}
\newblock In \emph{Proceedings of the 57th Conference of the Association for
  Computational Linguistics, {ACL} 2019, Florence, Italy, July 28- August 2,
  2019, Volume 1: Long Papers}, pages 4996--5001. Association for Computational
  Linguistics.

\bibitem[{Rama et~al.(2020)Rama, Beinborn, and
  Eger}]{DBLP:conf/coling/RamaBE20}
Taraka Rama, Lisa Beinborn, and Steffen Eger. 2020.
\newblock \href {https://doi.org/10.18653/v1/2020.coling-main.105} {Probing
  multilingual {BERT} for genetic and typological signals}.
\newblock In \emph{Proceedings of the 28th International Conference on
  Computational Linguistics, {COLING} 2020, Barcelona, Spain (Online), December
  8-13, 2020}, pages 1214--1228. International Committee on Computational
  Linguistics.

\bibitem[{Reimers and Gurevych(2019)}]{DBLP:conf/emnlp/ReimersG19}
Nils Reimers and Iryna Gurevych. 2019.
\newblock \href {https://doi.org/10.18653/v1/D19-1410} {Sentence-bert: Sentence
  embeddings using siamese bert-networks}.
\newblock In \emph{Proceedings of the 2019 Conference on Empirical Methods in
  Natural Language Processing and the 9th International Joint Conference on
  Natural Language Processing, {EMNLP-IJCNLP} 2019, Hong Kong, China, November
  3-7, 2019}, pages 3980--3990. Association for Computational Linguistics.

\bibitem[{Sabet et~al.(2020)Sabet, Dufter, Yvon, and
  Sch{\"{u}}tze}]{DBLP:conf/emnlp/SabetDYS20}
Masoud~Jalili Sabet, Philipp Dufter, Fran{\c{c}}ois Yvon, and Hinrich
  Sch{\"{u}}tze. 2020.
\newblock \href {https://doi.org/10.18653/v1/2020.findings-emnlp.147}
  {Simalign: High quality word alignments without parallel training data using
  static and contextualized embeddings}.
\newblock In \emph{Proceedings of the 2020 Conference on Empirical Methods in
  Natural Language Processing: Findings, {EMNLP} 2020, Online Event, 16-20
  November 2020}, volume {EMNLP} 2020 of \emph{Findings of {ACL}}, pages
  1627--1643. Association for Computational Linguistics.

\bibitem[{Saggion et~al.(2023)Saggion, {\v{S}}tajner, Ferr{\'e}s, Sheang,
  Shardlow, North, and Zampieri}]{saggion2023findings}
Horacio Saggion, Sanja {\v{S}}tajner, Daniel Ferr{\'e}s, Kim~Cheng Sheang,
  Matthew Shardlow, Kai North, and Marcos Zampieri. 2023.
\newblock Findings of the tsar-2022 shared task on multilingual lexical
  simplification.
\newblock \emph{arXiv preprint arXiv:2302.02888}.

\bibitem[{Scao et~al.(2022)Scao, Fan, Akiki, Pavlick, Ilic, Hesslow,
  Castagn{\'{e}}, Luccioni, Yvon, Gall{\'{e}}, Tow, Rush, Biderman, Webson,
  Ammanamanchi, Wang, Sagot, Muennighoff, del Moral, Ruwase, Bawden, Bekman,
  McMillan{-}Major, Beltagy, Nguyen, Saulnier, Tan, Suarez, Sanh,
  Lauren{\c{c}}on, Jernite, Launay, Mitchell, Raffel, Gokaslan, Simhi, Soroa,
  Aji, Alfassy, Rogers, Nitzav, Xu, Mou, Emezue, Klamm, Leong, van Strien,
  Adelani, and et~al.}]{DBLP:journals/corr/abs-2211-05100}
Teven~Le Scao, Angela Fan, Christopher Akiki, Ellie Pavlick, Suzana Ilic,
  Daniel Hesslow, Roman Castagn{\'{e}}, Alexandra~Sasha Luccioni,
  Fran{\c{c}}ois Yvon, Matthias Gall{\'{e}}, Jonathan Tow, Alexander~M. Rush,
  Stella Biderman, Albert Webson, Pawan~Sasanka Ammanamanchi, Thomas Wang,
  Beno{\^{\i}}t Sagot, Niklas Muennighoff, Albert~Villanova del Moral, Olatunji
  Ruwase, Rachel Bawden, Stas Bekman, Angelina McMillan{-}Major, Iz~Beltagy,
  Huu Nguyen, Lucile Saulnier, Samson Tan, Pedro~Ortiz Suarez, Victor Sanh,
  Hugo Lauren{\c{c}}on, Yacine Jernite, Julien Launay, Margaret Mitchell, Colin
  Raffel, Aaron Gokaslan, Adi Simhi, Aitor Soroa, Alham~Fikri Aji, Amit
  Alfassy, Anna Rogers, Ariel~Kreisberg Nitzav, Canwen Xu, Chenghao Mou, Chris
  Emezue, Christopher Klamm, Colin Leong, Daniel van Strien, David~Ifeoluwa
  Adelani, and et~al. 2022.
\newblock \href {https://doi.org/10.48550/arXiv.2211.05100} {{BLOOM:} {A}
  176b-parameter open-access multilingual language model}.
\newblock \emph{CoRR}, abs/2211.05100.

\bibitem[{Shliazhko et~al.(2022)Shliazhko, Fenogenova, Tikhonova, Mikhailov,
  Kozlova, and Shavrina}]{DBLP:journals/corr/abs-2204-07580}
Oleh Shliazhko, Alena Fenogenova, Maria Tikhonova, Vladislav Mikhailov,
  Anastasia Kozlova, and Tatiana Shavrina. 2022.
\newblock \href {https://doi.org/10.48550/arXiv.2204.07580} {mgpt: Few-shot
  learners go multilingual}.
\newblock \emph{CoRR}, abs/2204.07580.

\bibitem[{Tan et~al.(2019)Tan, Chen, He, Xia, Qin, and
  Liu}]{DBLP:conf/emnlp/TanCHXQL19}
Xu~Tan, Jiale Chen, Di~He, Yingce Xia, Tao Qin, and Tie{-}Yan Liu. 2019.
\newblock \href {https://doi.org/10.18653/v1/D19-1089} {Multilingual neural
  machine translation with language clustering}.
\newblock In \emph{Proceedings of the 2019 Conference on Empirical Methods in
  Natural Language Processing and the 9th International Joint Conference on
  Natural Language Processing, {EMNLP-IJCNLP} 2019, Hong Kong, China, November
  3-7, 2019}, pages 963--973. Association for Computational Linguistics.

\bibitem[{Tay et~al.(2022)Tay, Tran, Ruder, Gupta, Chung, Bahri, Qin,
  Baumgartner, Yu, and Metzler}]{DBLP:conf/iclr/Tay0RGCB0B0M22}
Yi~Tay, Vinh~Q. Tran, Sebastian Ruder, Jai~Prakash Gupta, Hyung~Won Chung, Dara
  Bahri, Zhen Qin, Simon Baumgartner, Cong Yu, and Donald Metzler. 2022.
\newblock \href {https://openreview.net/forum?id=JtBRnrlOEFN} {Charformer: Fast
  character transformers via gradient-based subword tokenization}.
\newblock In \emph{The Tenth International Conference on Learning
  Representations, {ICLR} 2022, Virtual Event, April 25-29, 2022}.
  OpenReview.net.

\bibitem[{Wang et~al.(2023)Wang, Adel, Lange, Str{\"{o}}tgen, and
  Sch{\"{u}}tze}]{DBLP:journals/corr/abs-2305-00090}
Mingyang Wang, Heike Adel, Lukas Lange, Jannik Str{\"{o}}tgen, and Hinrich
  Sch{\"{u}}tze. 2023.
\newblock \href {https://doi.org/10.48550/arXiv.2305.00090} {{NLNDE} at
  semeval-2023 task 12: Adaptive pretraining and source language selection for
  low-resource multilingual sentiment analysis}.
\newblock \emph{CoRR}, abs/2305.00090.

\bibitem[{Wu and Dredze(2019)}]{DBLP:conf/emnlp/WuD19}
Shijie Wu and Mark Dredze. 2019.
\newblock \href {https://doi.org/10.18653/v1/D19-1077} {Beto, bentz, becas: The
  surprising cross-lingual effectiveness of {BERT}}.
\newblock In \emph{Proceedings of the 2019 Conference on Empirical Methods in
  Natural Language Processing and the 9th International Joint Conference on
  Natural Language Processing, {EMNLP-IJCNLP} 2019, Hong Kong, China, November
  3-7, 2019}, pages 833--844. Association for Computational Linguistics.

\bibitem[{Xue et~al.(2022)Xue, Barua, Constant, Al{-}Rfou, Narang, Kale,
  Roberts, and Raffel}]{DBLP:journals/tacl/XueBCANKRR22}
Linting Xue, Aditya Barua, Noah Constant, Rami Al{-}Rfou, Sharan Narang, Mihir
  Kale, Adam Roberts, and Colin Raffel. 2022.
\newblock \href {https://doi.org/10.1162/tacl\_a\_00461} {Byt5: Towards a
  token-free future with pre-trained byte-to-byte models}.
\newblock \emph{Trans. Assoc. Comput. Linguistics}, 10:291--306.

\bibitem[{Xue et~al.(2021)Xue, Constant, Roberts, Kale, Al{-}Rfou, Siddhant,
  Barua, and Raffel}]{DBLP:conf/naacl/XueCRKASBR21}
Linting Xue, Noah Constant, Adam Roberts, Mihir Kale, Rami Al{-}Rfou, Aditya
  Siddhant, Aditya Barua, and Colin Raffel. 2021.
\newblock \href {https://doi.org/10.18653/v1/2021.naacl-main.41} {mt5: {A}
  massively multilingual pre-trained text-to-text transformer}.
\newblock In \emph{Proceedings of the 2021 Conference of the North American
  Chapter of the Association for Computational Linguistics: Human Language
  Technologies, {NAACL-HLT} 2021, Online, June 6-11, 2021}, pages 483--498.
  Association for Computational Linguistics.

\bibitem[{Ye et~al.(2023)Ye, Liu, and
  Sch{\"{u}}tze}]{DBLP:journals/corr/abs-2305-13401}
Haotian Ye, Yihong Liu, and Hinrich Sch{\"{u}}tze. 2023.
\newblock \href {https://doi.org/10.48550/ARXIV.2305.13401} {A study of
  conceptual language similarity: comparison and evaluation}.
\newblock \emph{CoRR}, abs/2305.13401.

\end{thebibliography}

\newpage

\appendix

\section{Languages}
\label{sec:languages}

Tab.~\ref{tab:language1}-\ref{tab:language6} show the language list covered by \mPLMs and corpora.

Tab.~\ref{tab:cross-lingual eval lang} provides the languages used for evaluating cross-lingual transfer.

\begin{table*}
    \centering
    \resizebox{0.95\textwidth}{!}{
\begin{tabular}{c|cccccccc|ccc}
    \toprule
     & \makecell{mBERT \\ CANINE-S \\ CANINE-C} & \makecell{XLM-R-Base \\ XLM-R-Large} & Glot500 & mGPT & mT5-Base & XLM-Align & NLLB-200 & XLS-R-300M & \Flores & \Bible & \Fleurs \\ \midrule
    ace\_Arab & & & & & & & \checkmark & & \checkmark & & \\
    ace\_Latn & & & \checkmark & & & & \checkmark & & \checkmark & \checkmark & \\
    ach\_Latn & & & \checkmark & & & & & & & \checkmark & \\
    acm\_Arab & & & \checkmark & & & & \checkmark & & \checkmark & & \\
    acq\_Arab & & & & & & & \checkmark & & \checkmark & & \\
    acr\_Latn & & & \checkmark & & & & & & & \checkmark & \\
    aeb\_Arab & & & & & & & \checkmark & & \checkmark & & \\
\bottomrule
    \end{tabular}
    }
    \caption{Languages covered by \mPLMs and corpora.}
    \label{tab:language1}
\end{table*}

\begin{table*}
    \centering
    \resizebox{0.95\textwidth}{!}{
    \begin{tabular}{c|cccccccc|ccc}
    \toprule
         & \makecell{mBERT \\ CANINE-S \\ CANINE-C} & \makecell{XLM-R-Base \\ XLM-R-Large} & Glot500 & mGPT & mT5-Base & XLM-Align & NLLB-200 & XLS-R-300M & \Flores & \Bible & \Fleurs \\ \midrule
        ckb\_Arab & & \checkmark & \checkmark & & \checkmark & \checkmark & \checkmark & \checkmark & \checkmark & \checkmark & \checkmark \\
        ckb\_Latn & & & \checkmark & & & & & & & \checkmark & \\
        cmn\_Hani & \checkmark & \checkmark & \checkmark & \checkmark & \checkmark & \checkmark & \checkmark & \checkmark & \checkmark & \checkmark & \checkmark \\
        cnh\_Latn & & & \checkmark & & & & & \checkmark & & \checkmark & \\
        crh\_Cyrl & & & \checkmark & & & & & & & \checkmark & \\
        crh\_Latn & & & \checkmark & & & & \checkmark & & \checkmark & & \\
        crs\_Latn & & & \checkmark & & & & & & & \checkmark & \\
\bottomrule
    \end{tabular}
    }
    \caption{Languages covered by \mPLMs and corpora.}
    \label{tab:language2}
\end{table*}

\begin{table*}
    \centering
    \resizebox{0.95\textwidth}{!}{
    \begin{tabular}{c|cccccccc|ccc}
    \toprule
         & \makecell{mBERT \\ CANINE-S \\ CANINE-C} & \makecell{XLM-R-Base \\ XLM-R-Large} & Glot500 & mGPT & mT5-Base & XLM-Align & NLLB-200 & XLS-R-300M & \Flores & \Bible & \Fleurs \\ \midrule
        hus\_Latn & & & \checkmark & & & & & & & \checkmark & \\
        hye\_Armn & \checkmark & \checkmark & \checkmark & \checkmark & \checkmark & \checkmark & \checkmark & \checkmark & \checkmark & \checkmark & \checkmark \\
        iba\_Latn & & & \checkmark & & & & & & & \checkmark & \\
        ibo\_Latn & & & \checkmark & & \checkmark & & \checkmark & & \checkmark & \checkmark & \checkmark \\
        ifa\_Latn & & & \checkmark & & & & & & & \checkmark & \\
        ifb\_Latn & & & \checkmark & & & & & & & \checkmark & \\
        ikk\_Latn & & & \checkmark & & & & & & & \checkmark & \\
\bottomrule
    \end{tabular}
    }
    \caption{Languages covered by \mPLMs and corpora.}
    \label{tab:language3}
\end{table*}

\begin{table*}
    \centering
    \resizebox{0.95\textwidth}{!}{
    \begin{tabular}{c|cccccccc|ccc}
    \toprule
         & \makecell{mBERT \\ CANINE-S \\ CANINE-C} & \makecell{XLM-R-Base \\ XLM-R-Large} & Glot500 & mGPT & mT5-Base & XLM-Align & NLLB-200 & XLS-R-300M & \Flores & \Bible & \Fleurs \\ \midrule
        luo\_Latn & & & \checkmark & & & & \checkmark & & \checkmark & \checkmark & \\
        lus\_Latn & & & \checkmark & & & & \checkmark & & \checkmark & \checkmark & \\
        lvs\_Latn & & & \checkmark & & & & \checkmark & & \checkmark & & \\
        lzh\_Hani & & & \checkmark & & & & & & & \checkmark & \\
        mad\_Latn & & & \checkmark & & & & & & & \checkmark & \\
        mag\_Deva & & & & & & & \checkmark & & \checkmark & & \\
        mah\_Latn & & & \checkmark & & & & & & & \checkmark & \\
\bottomrule
    \end{tabular}
    }
    \caption{Languages covered by \mPLMs and corpora.}
    \label{tab:language4}
\end{table*}

\begin{table*}
    \centering
    \resizebox{0.95\textwidth}{!}{
    \begin{tabular}{c|cccccccc|ccc}
    \toprule
         & \makecell{mBERT \\ CANINE-S \\ CANINE-C} & \makecell{XLM-R-Base \\ XLM-R-Large} & Glot500 & mGPT & mT5-Base & XLM-Align & NLLB-200 & XLS-R-300M & \Flores & \Bible & \Fleurs \\ \midrule
        pap\_Latn & & & \checkmark & & & & \checkmark & & \checkmark & \checkmark & \\
        pau\_Latn & & & \checkmark & & & & & & & \checkmark & \\
        pbt\_Arab & & & & & & & \checkmark & & \checkmark & & \\
        pcm\_Latn & & & \checkmark & & & & & & & \checkmark & \\
        pdt\_Latn & & & \checkmark & & & & & & & \checkmark & \\
        pes\_Arab & \checkmark & \checkmark & \checkmark & \checkmark & \checkmark & \checkmark & \checkmark & \checkmark & \checkmark & \checkmark & \\
        pis\_Latn & & & \checkmark & & & & & & & \checkmark & \\
\bottomrule
    \end{tabular}
    }
    \caption{Languages covered by \mPLMs and corpora.}
    \label{tab:language5}
\end{table*}

\begin{table*}
    \centering
    \resizebox{0.95\textwidth}{!}{
    \begin{tabular}{c|cccccccc|ccc}
    \toprule
         & \makecell{mBERT \\ CANINE-S \\ CANINE-C} & \makecell{XLM-R-Base \\ XLM-R-Large} & Glot500 & mGPT & mT5-Base & XLM-Align & NLLB-200 & XLS-R-300M & \Flores & \Bible & \Fleurs \\ \midrule
        tdt\_Latn & & & \checkmark & & & & & & & \checkmark & \\
        tel\_Telu & \checkmark & \checkmark & \checkmark & \checkmark & \checkmark & \checkmark & \checkmark & \checkmark & \checkmark & \checkmark & \\
        teo\_Latn & & & \checkmark & & & & & & & \checkmark & \\
        tgk\_Cyrl & \checkmark & & \checkmark & \checkmark & \checkmark & \checkmark & \checkmark & \checkmark & \checkmark & \checkmark & \\
        tgl\_Latn & \checkmark & \checkmark & \checkmark & \checkmark & & \checkmark & \checkmark & \checkmark & \checkmark & \checkmark & \\
        tha\_Thai & & \checkmark & \checkmark & \checkmark & \checkmark & \checkmark & \checkmark & \checkmark & \checkmark & \checkmark & \checkmark \\
        tih\_Latn & & & \checkmark & & & & & & & \checkmark & \\
\bottomrule
    \end{tabular}
    }
    \caption{Languages covered by \mPLMs and corpora.}
    \label{tab:language6}
\end{table*}

\begin{table*}
    \centering
    \resizebox{0.95\textwidth}{!}{
    \begin{tabular}{cccccc}
    \toprule
    \multicolumn{1}{c|}{Task} & \multicolumn{5}{c}{Language List} \\
    \midrule
    \multicolumn{1}{c|}{NER (108)} & \multicolumn{5}{l}{\makecell{ace\_Latn, afr\_Latn, als\_Latn, amh\_Ethi, arz\_Arab, asm\_Beng, ast\_Latn, azj\_Latn, bak\_Cyrl, bel\_Cyrl, ben\_Beng, bho\_Deva, bod\_Tibt, bos\_Latn, bul\_Cyrl, \\cat\_Latn, ceb\_Latn, ces\_Latn, ckb\_Arab, crh\_Latn, cym\_Latn, dan\_Latn, deu\_Latn, ekk\_Latn, ell\_Grek, epo\_Latn, eus\_Latn, fao\_Latn, fin\_Latn, fra\_Latn, \\fur\_Latn, gla\_Latn, gle\_Latn, glg\_Latn, gug\_Latn, guj\_Gujr, heb\_Hebr, hrv\_Latn, hun\_Latn, hye\_Armn, ibo\_Latn, ilo\_Latn, ind\_Latn, isl\_Latn, ita\_Latn, \\jav\_Latn, jpn\_Jpan, kan\_Knda, kat\_Geor, kaz\_Cyrl, khm\_Khmr, kin\_Latn, kir\_Cyrl, kor\_Hang, lij\_Latn, lim\_Latn, lin\_Latn, lit\_Latn, lmo\_Latn, ltz\_Latn, \\mal\_Mlym, mar\_Deva, min\_Latn, mkd\_Cyrl, mlt\_Latn, mri\_Latn, mya\_Mymr, nld\_Latn, nno\_Latn, oci\_Latn, ory\_Orya, pan\_Guru, pes\_Arab, plt\_Latn, pol\_Latn, \\por\_Latn, ron\_Latn, san\_Deva, scn\_Latn, sin\_Sinh, slk\_Latn, slv\_Latn, snd\_Arab, som\_Latn, srp\_Cyrl, sun\_Latn, swe\_Latn, swh\_Latn, szl\_Latn, tam\_Taml, \\tat\_Cyrl, tel\_Telu, tgk\_Cyrl, tgl\_Latn, tha\_Thai, tuk\_Latn, tur\_Latn, uig\_Arab, ukr\_Cyrl, urd\_Arab, uzn\_Latn, vec\_Latn, vie\_Latn, war\_Latn, ydd\_Hebr, \\yor\_Latn, yue\_Hani, zsm\_Latn}} \\
\bottomrule
    \end{tabular}
    }
    \caption{Languages for evaluating zero-shot cross-lingual transfer. The number in brackets is the number of the evaluated languages.}
    \label{tab:cross-lingual eval lang}
\end{table*}

\section{Comparison Across Corpora for \mPLMMeasure}
\label{sec:cmp_corpora}

\subsection{Monolingual vs. Parallel}
\label{sec:cmp_corpora1}

\begin{table}
    \centering 
    \resizebox{\columnwidth}{!}{
    \begin{tabular}{c|ccccc}
    \toprule
    & mPLM-Sim & Mono & 1 & 5 & 10 \\
    \midrule
LEX & 0.741 & 0.704 & 0.688 & 0.745 & 0.743 \\
GEN & 0.527 & 0.504 & 0.480 & 0.482 & 0.510 \\
GEO & 0.608 & 0.597 & 0.523 & 0.562 & 0.597 \\
SYN & 0.577 & 0.583 & 0.556 & 0.560 & 0.573 \\
INV & 0.248 & 0.245 & 0.226 & 0.265 & 0.260 \\
PHO & 0.094 & 0.109 & 0.114 & 0.118 & 0.102 \\
FEA & 0.358 & 0.369 & 0.347 & 0.371 & 0.360 \\
\midrule
AVG & \textbf{0.451} & 0.444 & 0.419 & 0.444 & 0.449 \\
    \bottomrule
    \end{tabular}
    }
    \caption{Comparison of pearson correlation result: Pearson correlation between seven similarity measurs and \mPLMMeasure (500 multi-parallel sentences), Mono (Monolingual corpora) and the results of using different amounts (1, 5, 10) of multi-parallel sentences.}
    \label{tab:cmp_pearson}
\end{table}

\begin{table}
    \centering 
    \resizebox{\columnwidth}{!}{
    \begin{tabular}{c|ccccc}
    \toprule
    & mPLM-Sim & Mono & 1 & 5 & 10 \\
    \midrule
NER & 0.647 & 0.644 & 0.644 & 0.646 & 0.647 \\
POS & 0.751 & 0.737 & 0.748 & 0.753 & 0.752 \\
Massive & 0.730 & 0.730 & 0.723 & 0.728 & 0.730 \\
Taxi & 0.583 & 0.585 & 0.580 & 0.582 & 0.582 \\
\midrule
AVG & \textbf{0.678} & 0.674 & 0.674 & 0.677 & \textbf{0.678}  \\
    \bottomrule
    \end{tabular}
    }
    \caption{Comparison of cross-lingual transfer result: Cross-lingual transfer result for four tasks from \mPLMMeasure (500 multi-parallel sentences), Mono (Monolingual corpora) and the results of using different amounts (1, 5, 10) of multi-parallel sentences.}
    \label{tab:cmp_xltransfer}
\end{table}

Both monolingual and parallel corpora can be exploited for obtaining sentence embeddings for measuring language similarity. We conduct experiments of exploiting monolingual corpora for measuring similarity across languages, and also provide the results of using different amounts (1, 5, 10, 500) of multi-parallel sentences.

For the experiment of pearson correlation in Sec. \ref{sec:correlation}, the results (MEAN) are shown in Tab. \ref{tab:cmp_pearson}.
For the experiment of cross-lingual transfer in Sec. \ref{sec:cross-lingual transfer}, the results are shown in Tab. \ref{tab:cmp_xltransfer}.
Based on these two experiments, we have the conclusions below:
\begin{itemize}
    \item mPLM-Sim using multi-parallel corpora achieves slightly better results than using monolingual corpora.
    \item mPLM-Sim (500 sentences) requires less data than exploiting monolingual corpora. Besides, using mPLM-Sim (10 sentences) can achieve comparable results with mPLM-Sim (500 sentences). While including a truly low-resource language for similarity measurement, mPLM-Sim requires around 10 sentences parallel to one existing language, while monolingual corpora requires massive sentences.
\end{itemize}

In a word, exploiting parallel corpora is better for measuring language similarity than monolingual corpora.

\subsection{\Flores vs. \Bible}
\label{sec:cmp_corpora2}

\begin{table}
    \centering 
    \resizebox{0.75\columnwidth}{!}{
    \begin{tabular}{c|cc|cc}
    \toprule
    & \multicolumn{2}{c|}{\Flores} & \multicolumn{2}{c}{\Bible} \\ \cline{2-5}
    & M & Mdn & M & Mdn \\
    \midrule
    LEX & 0.741& 0.864& 0.654 & 0.735 \\
    GEN & 0.527& 0.600& 0.519 & 0.572 \\
    GEO & 0.608& 0.674& 0.546 & 0.603 \\
    SYN & 0.577& 0.607& 0.491 & 0.528 \\
    INV & 0.248& 0.293& 0.254 & 0.276 \\
    PHO & 0.094& 0.144& 0.103 & 0.098 \\
    FEA & 0.358& 0.372& 0.333 & 0.357 \\ \midrule
    AVG & 0.451& 0.508& 0.414 & 0.453 \\
    \bottomrule
    \end{tabular}
    }
    \caption{Comparison across corpora: Pearson correlation between \mPLMMeasure and linguistic similarity measures for Glot500 and all corpora on 32 languages. \Flores achieves higher correlations than \Bible.}
    \label{tab:corpora}
\end{table}

To investigate the impact of multi-parallel corpora on the performance of \mPLMMeasure, we compare the results of Glot500 with \Flores and \Bible on 32 languages that are covered by both corpora.

Tab.~\ref{tab:corpora} shows that \Flores outperforms \Bible across all similarity measures, except for PHO. To gain further insights, we conduct a case study focusing on languages that exhibit different performances between the two corpora.

In comparison to \Bible, \Flores  consists of text that is closer to web content and spans a wider range of general domains. For example, a significant portion of Arabic script in \Flores is written without short vowels, which are commonly used in texts requiring strict adherence to precise pronunciation, such as the Bible.\footnote{\url{https://en.wikipedia.org/wiki/Arabic_diacritics}} This discrepancy leads to challenges in tokenization and representation for languages written in Arabic, such as Moroccan Arabic (ary\_Arab) and Egyptian Arabic (arz\_Arab), resulting in poorer performance.

\section{Visualization and Analysis Across Layers}
\label{sec:va_layer}

\subsection{Hierarchical Clustering Analysis}
\label{sec:hca}

\begin{figure*}
    \centering
    \begin{subfigure}[t]{0.475\textwidth}
        \centering
        \resizebox{\textwidth}{!}{
            \includegraphics[trim={0.1cm 0.1cm 0.1cm 0.1cm},clip]{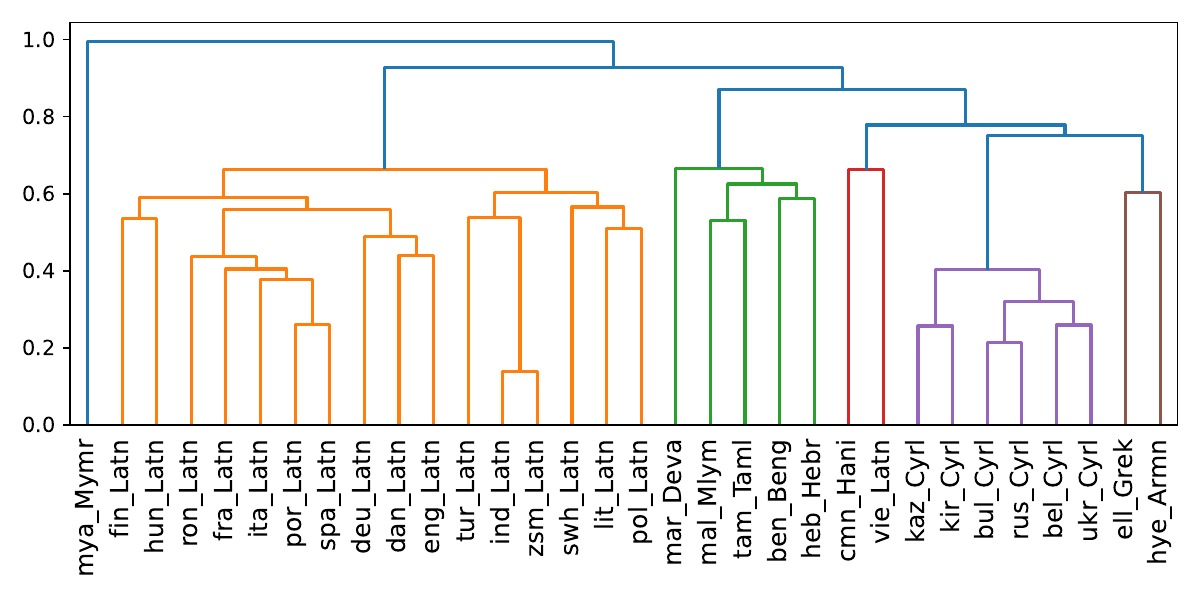}
        }
        \caption{Layer 0}
    \end{subfigure}%
    ~ 
    \begin{subfigure}[t]{0.475\textwidth}
        \centering
        \resizebox{\textwidth}{!}{
            \includegraphics[trim={0.1cm 0.1cm 0.1cm 0.1cm},clip]{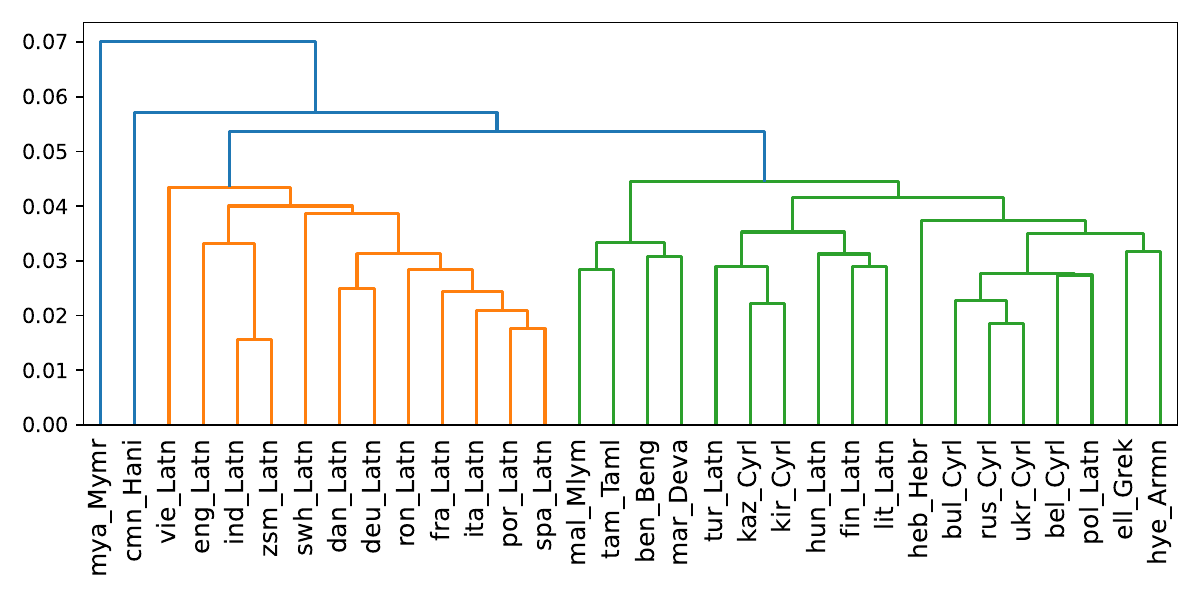}
        }
        \caption{Layer 4}
    \end{subfigure}
    ~ 
    \begin{subfigure}[t]{0.475\textwidth}
        \centering
        \resizebox{\textwidth}{!}{
            \includegraphics[trim={0.1cm 0.1cm 0.1cm 0.1cm},clip]{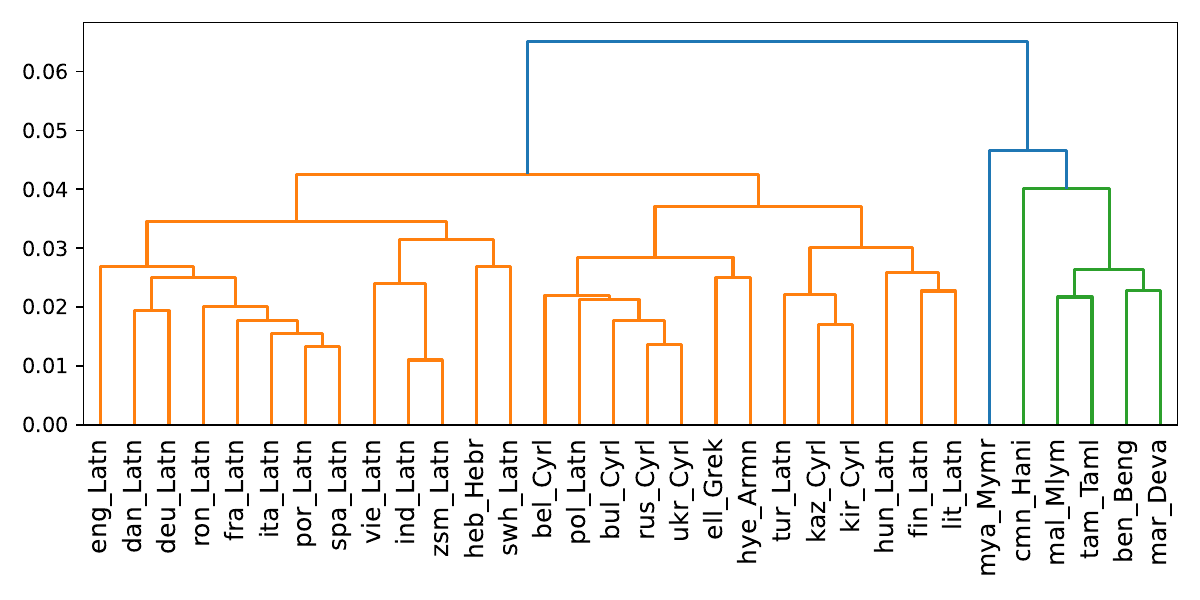}
        }
        \caption{Layer 8}
    \end{subfigure}
    ~
    \begin{subfigure}[t]{0.475\textwidth}
        \centering
        \resizebox{\textwidth}{!}{
            \includegraphics[trim={0.1cm 0.1cm 0.1cm 0.1cm},clip]{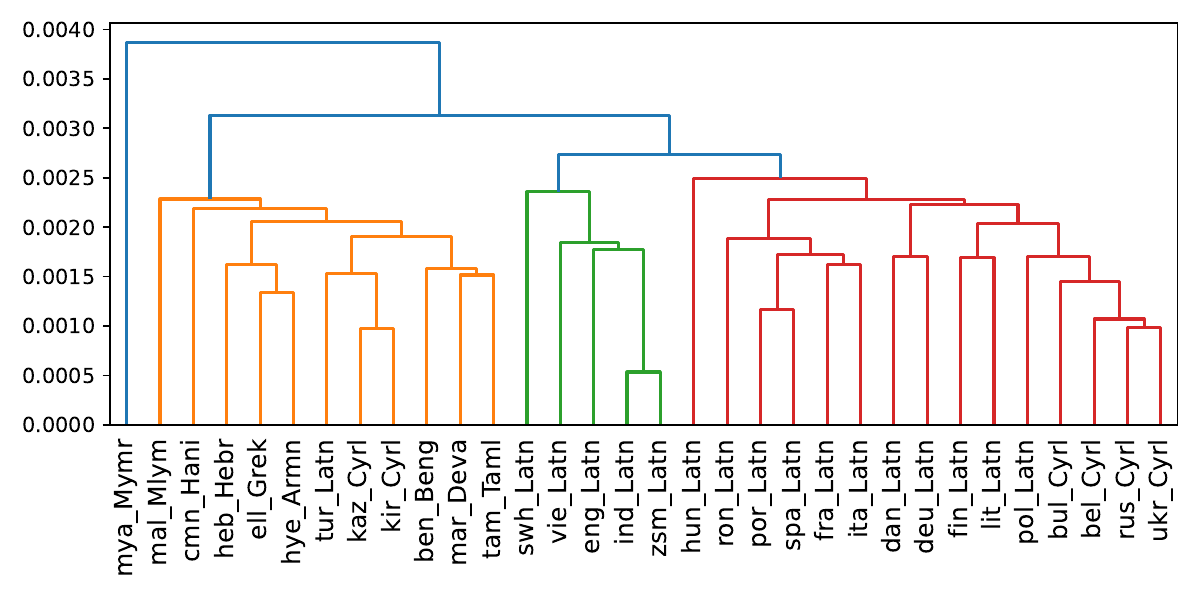}
        }
        \caption{Layer 12}
    \end{subfigure}
    \caption{Dendrograms illustrating hierarchical clustering results at layer 0, 4, 8, and 12 for Glot500 and \Flores across 32 languages.}
    \label{fig:dendrogram}
\end{figure*}

We conducted hierarchical
clustering analysis at different layers (0, 4, 8, and 12)
using the setting of Glot500 and \Flores
for \mPLMMeasure. The results, shown in
Fig.~\ref{fig:dendrogram}, reveal distinct patterns of
language clustering. In layer 0, the clustering primarily
emphasizes lexical similarities, with languages sharing the
same scripts being grouped together. As we progress to
layers 4 and 8, more high-level similarity patterns beyond
the surface-level are captured. For instance in these
layers, Turkish (tur\_Latn) and Polish (pol\_Latn) are
clustered with their Turkic and Slavic relatives although
they use different writing systems.
The similarity results of layer 12 are comparatively worse than those of the middle layers. For instance, English (eng\_Latn) deviates from its Germanic and Indo-European relatives and instead clusters with Malay languages (ind\_Latn, zsm\_Latn). This phenomenon can be attributed to the higher layer exhibiting lower inter-cluster distances (comparison between the y-axis range across figures of different layers), which diminishes its ability to effectively discriminate between language clusters.

\subsection{Similarity Heatmaps}
\label{sec:heatmap}

\begin{figure*}
  \centering
  \resizebox {0.8\textwidth} {!} {
    \includegraphics[trim={0cm 0cm 2cm 0cm},clip]{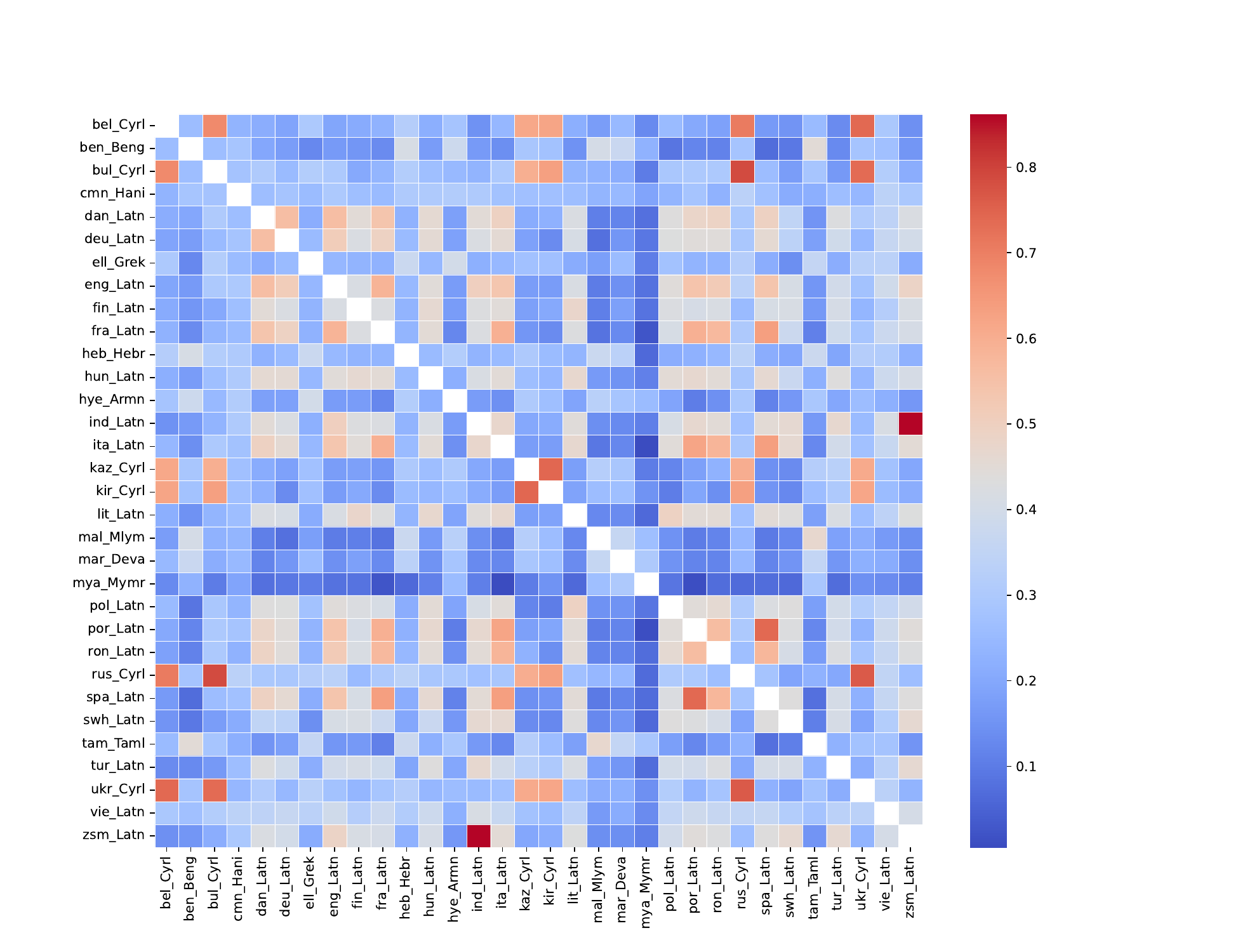}
  }
  \caption{Heatmaps of cosine similarity results at layer 0 for Glot500 and \Flores across 32 languages.}
  \label{fig:heatmap1}
\end{figure*}

\begin{figure*}
  \centering
  \resizebox {0.8\textwidth} {!} {
    \includegraphics[trim={0cm 0cm 2cm 0cm},clip]{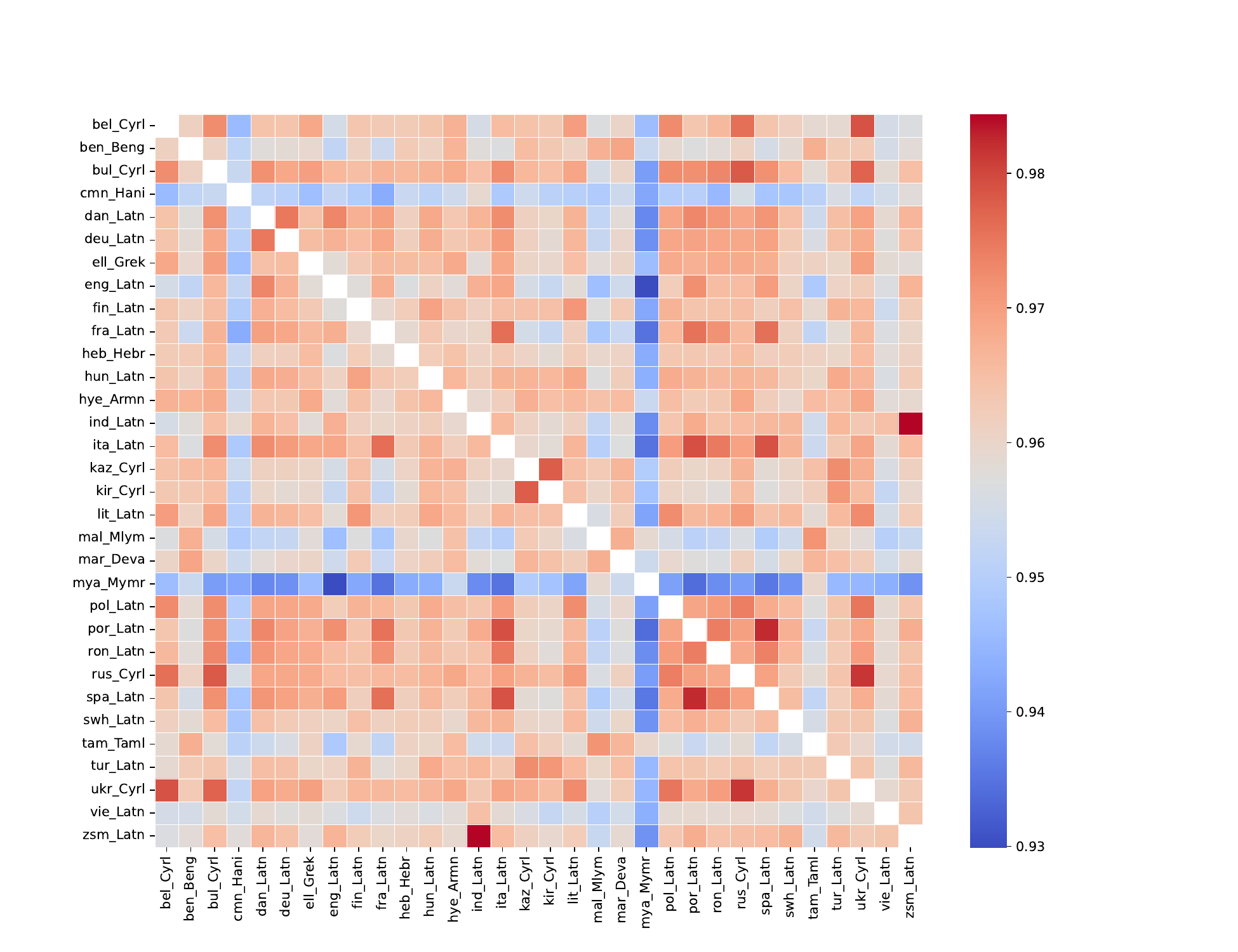}
  }
  \caption{Heatmaps of cosine similarity results at layer 4 for Glot500 and \Flores across 32 languages.}
  \label{fig:heatmap2}
\end{figure*}

\begin{figure*}
  \centering
  \resizebox {0.8\textwidth} {!} {
    \includegraphics[trim={0cm 0cm 2cm 0cm},clip]{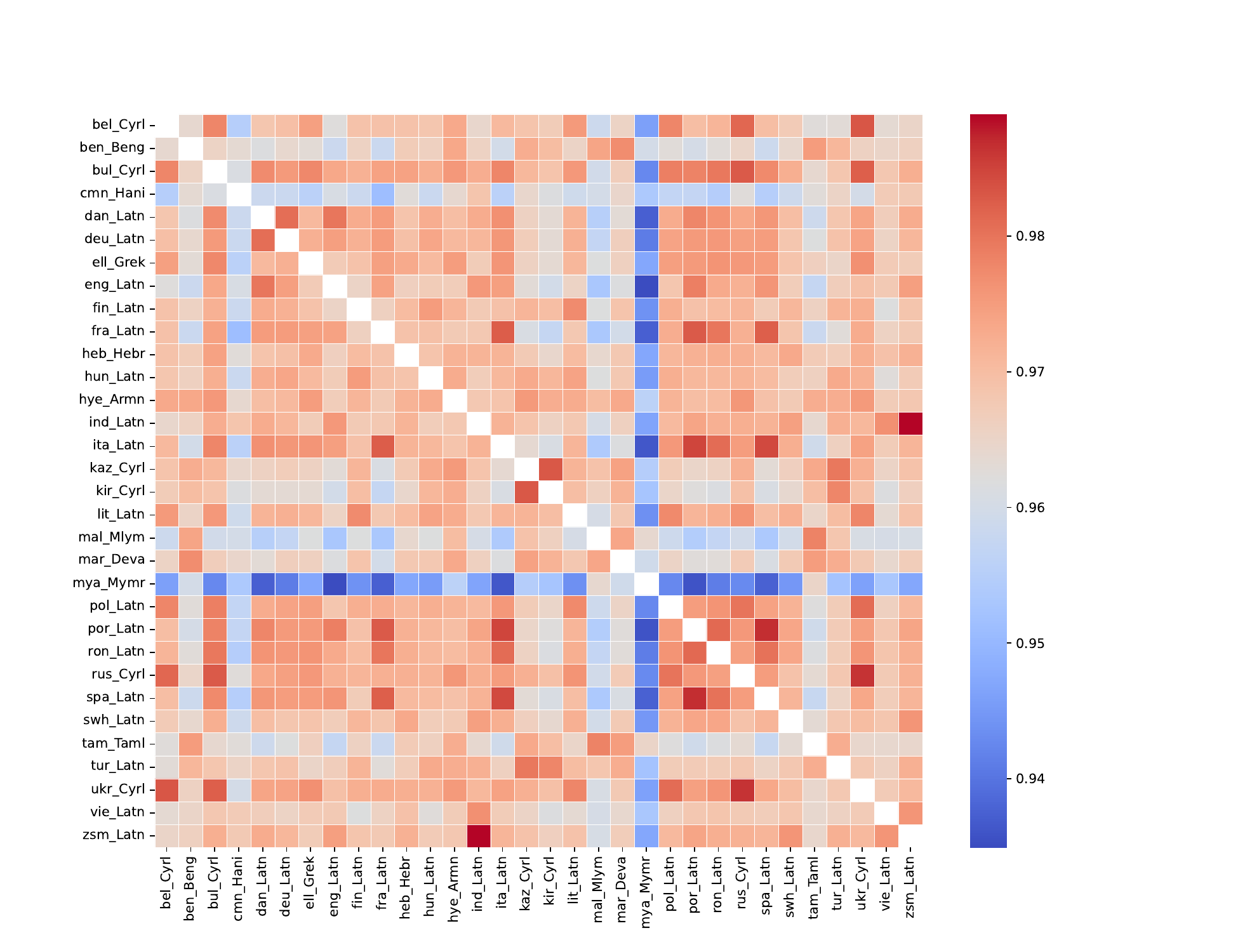}
  }
  \caption{Heatmaps of cosine similarity results at layer 8 for Glot500 and \Flores across 32 languages.}
  \label{fig:heatmap3}
\end{figure*}

\begin{figure*}
  \centering
  \resizebox {0.8\textwidth} {!} {
    \includegraphics[trim={0cm 0cm 2cm 0cm},clip]{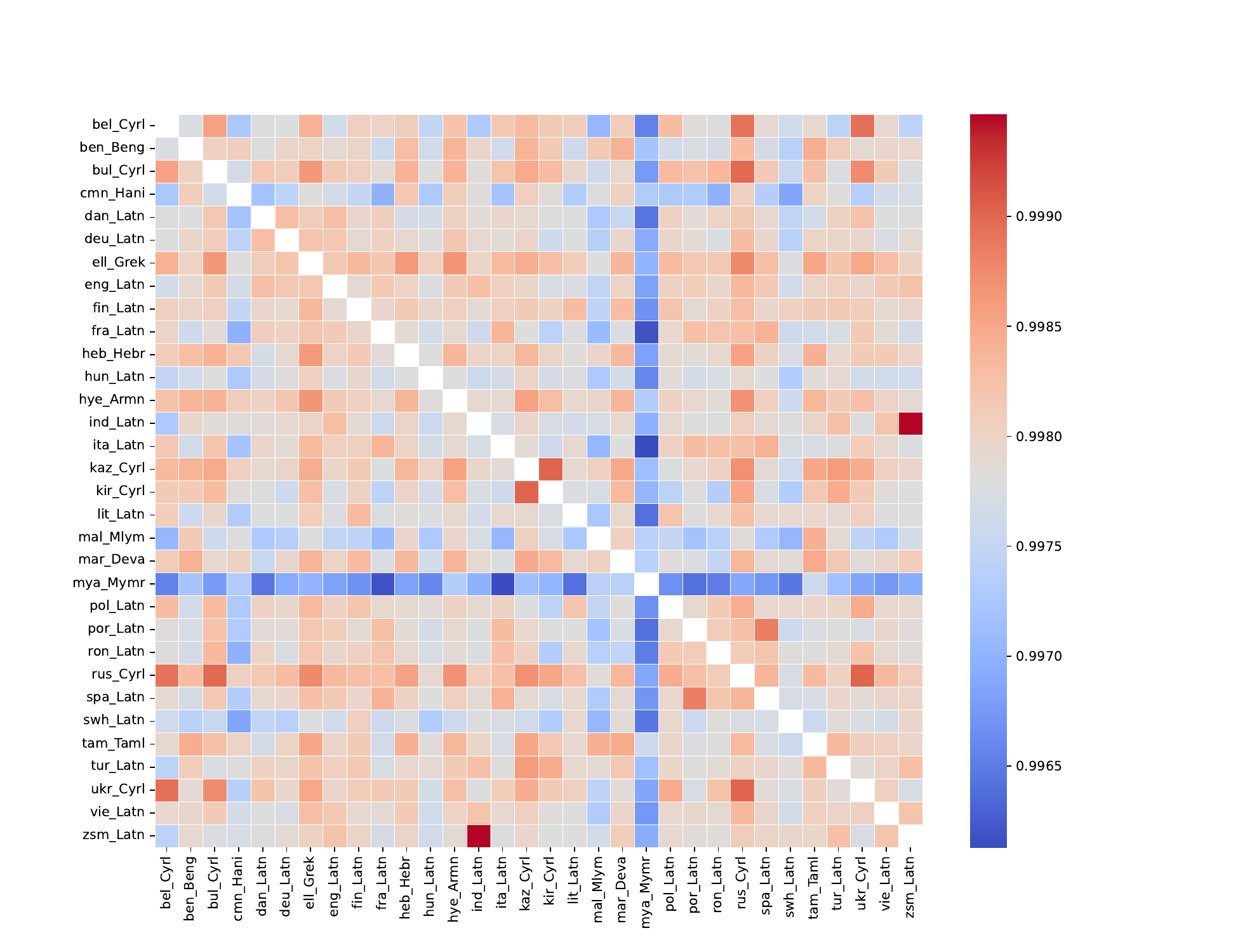}
  }
  \caption{Heatmaps of cosine similarity results at layer 12 for Glot500 and \Flores across 32 languages.}
  \label{fig:heatmap4}
\end{figure*}

Fig. \ref{fig:heatmap1}-\ref{fig:heatmap4} show the cosine simlarity values in heatmaps at layer 0, 4, 8 and 12, using the Glot500 and \Flores settings for \mPLMMeasure. 

Generally, as the layer number increases, higher cosine similarity values are observed. Layer 0 exhibits a significant contrast in similarity values, whereas layer 12 demonstrates very low contrast. Notably, Burmese (mya\_Mymr) consistently receives the lowest values across all layers, indicating the relationship between Burmese and other languages may be not well modeled.

\section{Analysis on Unseen Languages of mPLMs}
\label{sec:unseen}

The success of mPLM-Sim depends on the supporting languages of mPLMs. To get more insights about languages which are this not supported by a specific mPLM, we conduct a new Pearson correlation experiment based on 94 languages unseen by XLM-R. Among 94 languages, there are 24 (25.5\%) languages that achieve higher correlation than the average level of seen languages. These 24 languages usually have close languages seen by XLM-R, e.g, the unseen language, Cantonese (yue\_Hani) is close to Mandarin (cmn\_Hani). It shows that mPLM-Sim can be directly applied to some unseen languages which have close seen languages.

For the unseen languages which mPLM-Sim performs poorly, we can connect it to seen languages using traditional linguistic features, e.g., language family, and then use or weight the similarity results of seen languages as the results of the unseen languages. Since it is shown that mPLM-Sim provides better results than traditional linguistic features in our paper, connecting unseen languages to seen languages would be beneficial for unseen languages.

\section{Detailed Results of Cross-Lingual Transfer}
\label{sec:results}

We report the detailed results for all tasks and languages in Tab. \ref{tab:ner1}-\ref{tab:ner2} (NER), \ref{tab:pos1} (POS), \ref{tab:massive} (MASSIVE), \ref{tab:taxi1}-\ref{tab:taxi3} (Taxi1500).

\begin{table*}
    \centering
    \resizebox{0.925\textwidth}{!}{
\begin{tabular}{c|c|cc|cc|cc|cc|cc}
    \toprule
& ENG & \multicolumn{2}{c|}{LEX} & \multicolumn{2}{c|}{GEN} & \multicolumn{2}{c|}{GEO} & \multicolumn{2}{c|}{FEA} & \multicolumn{2}{c}{mPLM-Sim} \\
\midrule
ace\_Latn & 0.421 & 0.421 & eng\_Latn & 0.421 & eng\_Latn & 0.427 & hin\_Deva & 0.421 & eng\_Latn & \textbf{0.439} & spa\_Latn \\
afr\_Latn & \textbf{0.739} & \textbf{0.739} & eng\_Latn & \textbf{0.739} & eng\_Latn & 0.720 & arb\_Arab & 0.707 & rus\_Cyrl & \textbf{0.739} & eng\_Latn \\
als\_Latn & 0.767 & 0.767 & eng\_Latn & 0.737 & rus\_Cyrl & \textbf{0.774} & spa\_Latn & 0.737 & rus\_Cyrl & \textbf{0.774} & spa\_Latn \\
amh\_Ethi & 0.450 & 0.389 & cmn\_Hani & 0.515 & arb\_Arab & 0.515 & arb\_Arab & \textbf{0.554} & hin\_Deva & \textbf{0.554} & hin\_Deva \\
arz\_Arab & 0.491 & \textbf{0.715} & arb\_Arab & \textbf{0.715} & arb\_Arab & \textbf{0.715} & arb\_Arab & 0.491 & eng\_Latn & \textbf{0.715} & arb\_Arab \\
asm\_Beng & 0.661 & 0.603 & arb\_Arab & \textbf{0.720} & hin\_Deva & \textbf{0.720} & hin\_Deva & \textbf{0.720} & hin\_Deva & \textbf{0.720} & hin\_Deva \\
ast\_Latn & 0.813 & \textbf{0.857} & spa\_Latn & \textbf{0.857} & spa\_Latn & \textbf{0.857} & spa\_Latn & 0.680 & hin\_Deva & \textbf{0.857} & spa\_Latn \\
\bottomrule
    \end{tabular}
    }
    \caption{Cross-Lingual Transfer Results of NER (Part 1): The first column is the target language. For each language similarity measure, we report both the source language selected based on similarity and also the evaluation results on target language using the source language. For \mPLMMeasure, we report the layer achieving best performance (layer 1).}
    \label{tab:ner1}
\end{table*}

\begin{table*}
    \centering
    \resizebox{0.925\textwidth}{!}{
\begin{tabular}{c|c|cc|cc|cc|cc|cc}
    \toprule
& ENG & \multicolumn{2}{c|}{LEX} & \multicolumn{2}{c|}{GEN} & \multicolumn{2}{c|}{GEO} & \multicolumn{2}{c|}{FEA} & \multicolumn{2}{c}{mPLM-Sim} \\
\midrule
lij\_Latn & \textbf{0.431} & \textbf{0.431} & eng\_Latn & 0.413 & spa\_Latn & 0.413 & spa\_Latn & 0.395 & hin\_Deva & 0.413 & spa\_Latn \\
lim\_Latn & \textbf{0.646} & \textbf{0.646} & eng\_Latn & \textbf{0.646} & eng\_Latn & \textbf{0.646} & eng\_Latn & 0.605 & hin\_Deva & 0.621 & spa\_Latn \\
lin\_Latn & 0.486 & 0.486 & eng\_Latn & 0.486 & eng\_Latn & \textbf{0.555} & arb\_Arab & 0.486 & eng\_Latn & 0.519 & spa\_Latn \\
lit\_Latn & \textbf{0.707} & \textbf{0.707} & eng\_Latn & 0.699 & rus\_Cyrl & \textbf{0.707} & eng\_Latn & 0.699 & rus\_Cyrl & 0.699 & rus\_Cyrl \\
lmo\_Latn & \textbf{0.712} & \textbf{0.712} & eng\_Latn & 0.706 & spa\_Latn & 0.706 & spa\_Latn & 0.559 & hin\_Deva & 0.706 & spa\_Latn \\
ltz\_Latn & 0.646 & 0.646 & eng\_Latn & 0.646 & eng\_Latn & 0.646 & eng\_Latn & \textbf{0.663} & spa\_Latn & \textbf{0.663} & spa\_Latn \\
mal\_Mlym & 0.591 & 0.642 & arb\_Arab & 0.591 & eng\_Latn & \textbf{0.709} & hin\_Deva & \textbf{0.709} & hin\_Deva & \textbf{0.709} & hin\_Deva \\
\bottomrule
    \end{tabular}
    }
    \caption{Cross-Lingual Transfer Results of NER (Part 2): The first column is the target language. For each language similarity measure, we report both the source language selected based on similarity and also the evaluation results on target language using the source language. For \mPLMMeasure, we report the layer achieving best performance (layer 1).}
    \label{tab:ner2}
\end{table*}

\begin{table*}
    \centering
    \resizebox{0.925\textwidth}{!}{
\begin{tabular}{c|c|cc|cc|cc|cc|cc}
    \toprule
& ENG & \multicolumn{2}{c|}{LEX} & \multicolumn{2}{c|}{GEN} & \multicolumn{2}{c|}{GEO} & \multicolumn{2}{c|}{FEA} & \multicolumn{2}{c}{mPLM-Sim} \\
\midrule
afr\_Latn & 0.850 & 0.850 & eng\_Latn & 0.850 & eng\_Latn & 0.599 & arb\_Arab & 0.809 & rus\_Cyrl & \textbf{0.854} & spa\_Latn \\
ajp\_Arab & \textbf{0.671} & 0.648 & arb\_Arab & 0.648 & arb\_Arab & 0.648 & arb\_Arab & 0.651 & hin\_Deva & 0.648 & arb\_Arab \\
amh\_Ethi & 0.648 & 0.645 & cmn\_Hani & 0.670 & arb\_Arab & 0.670 & arb\_Arab & \textbf{0.704} & hin\_Deva & \textbf{0.704} & hin\_Deva \\
bam\_Latn & 0.451 & 0.451 & eng\_Latn & 0.451 & eng\_Latn & 0.411 & spa\_Latn & \textbf{0.484} & hin\_Deva & 0.411 & spa\_Latn \\
bel\_Cyrl & 0.824 & \textbf{0.934} & rus\_Cyrl & \textbf{0.934} & rus\_Cyrl & 0.824 & eng\_Latn & 0.719 & arb\_Arab & \textbf{0.934} & rus\_Cyrl \\
ben\_Beng & 0.767 & 0.583 & arb\_Arab & \textbf{0.803} & hin\_Deva & \textbf{0.803} & hin\_Deva & \textbf{0.803} & hin\_Deva & \textbf{0.803} & hin\_Deva \\
bho\_Deva & 0.520 & \textbf{0.682} & hin\_Deva & \textbf{0.682} & hin\_Deva & \textbf{0.682} & hin\_Deva & 0.536 & arb\_Arab & \textbf{0.682} & hin\_Deva \\
\bottomrule
    \end{tabular}
    }
    \caption{Cross-Lingual Transfer Results of POS: The first column is the target language. For each language similarity measure, we report both the source language selected based on similarity and also the evaluation results on target language using the source language. For \mPLMMeasure, we report the layer achieving best performance (layer 2).}
    \label{tab:pos1}
\end{table*}

\begin{table*}
    \centering
    \resizebox{0.925\textwidth}{!}{
\begin{tabular}{c|c|cc|cc|cc|cc|cc}
    \toprule
& ENG & \multicolumn{2}{c|}{LEX} & \multicolumn{2}{c|}{GEN} & \multicolumn{2}{c|}{GEO} & \multicolumn{2}{c|}{FEA} & \multicolumn{2}{c}{mPLM-Sim} \\
\midrule
afr\_Latn & \textbf{0.732} & \textbf{0.732} & eng\_Latn & \textbf{0.732} & eng\_Latn & 0.589 & arb\_Arab & 0.701 & rus\_Cyrl & \textbf{0.732} & eng\_Latn \\
als\_Latn & 0.708 & 0.708 & eng\_Latn & 0.721 & rus\_Cyrl & \textbf{0.727} & spa\_Latn & \textbf{0.727} & spa\_Latn & \textbf{0.727} & spa\_Latn \\
amh\_Ethi & 0.557 & 0.470 & cmn\_Hani & 0.532 & arb\_Arab & 0.532 & arb\_Arab & \textbf{0.611} & hin\_Deva & \textbf{0.611} & hin\_Deva \\
azj\_Latn & 0.773 & 0.773 & eng\_Latn & 0.773 & eng\_Latn & 0.705 & arb\_Arab & \textbf{0.793} & hin\_Deva & \textbf{0.793} & hin\_Deva \\
ben\_Beng & 0.676 & 0.625 & arb\_Arab & \textbf{0.768} & hin\_Deva & \textbf{0.768} & hin\_Deva & \textbf{0.768} & hin\_Deva & \textbf{0.768} & hin\_Deva \\
cat\_Latn & 0.731 & \textbf{0.833} & spa\_Latn & \textbf{0.833} & spa\_Latn & \textbf{0.833} & spa\_Latn & 0.731 & eng\_Latn & \textbf{0.833} & spa\_Latn \\
cym\_Latn & 0.492 & 0.492 & eng\_Latn & \textbf{0.495} & rus\_Cyrl & 0.492 & eng\_Latn & 0.433 & arb\_Arab & 0.480 & spa\_Latn \\
\bottomrule
\end{tabular}

    }
    \caption{Cross-Lingual Transfer Result of MASSIVE: The first column is the target language. For each language similarity measure, we report both the source language selected based on similarity and also the evaluation results on target language using the source language. For \mPLMMeasure, we report the layer achieving best performance (layer 8).}
    \label{tab:massive}
\end{table*}

\begin{table*}
    \centering
    \resizebox{0.925\textwidth}{!}{
\begin{tabular}{c|c|cc|cc|cc|cc|cc}
    \toprule
& ENG & \multicolumn{2}{c|}{LEX} & \multicolumn{2}{c|}{GEN} & \multicolumn{2}{c|}{GEO} & \multicolumn{2}{c|}{FEA} & \multicolumn{2}{c}{mPLM-Sim} \\
\midrule
ace\_Latn & 0.624 & 0.624 & eng\_Latn & 0.624 & eng\_Latn & \textbf{0.726} & hin\_Deva & 0.624 & eng\_Latn & 0.654 & spa\_Latn \\
afr\_Latn & 0.600 & 0.600 & eng\_Latn & 0.600 & eng\_Latn & 0.455 & arb\_Arab & 0.522 & rus\_Cyrl & \textbf{0.604} & spa\_Latn \\
aka\_Latn & \textbf{0.518} & \textbf{0.518} & eng\_Latn & \textbf{0.518} & eng\_Latn & 0.471 & spa\_Latn & 0.469 & hin\_Deva & 0.471 & spa\_Latn \\
als\_Latn & \textbf{0.575} & \textbf{0.575} & eng\_Latn & 0.557 & rus\_Cyrl & 0.536 & spa\_Latn & 0.557 & rus\_Cyrl & 0.536 & spa\_Latn \\
ary\_Arab & 0.421 & \textbf{0.484} & arb\_Arab & \textbf{0.484} & arb\_Arab & 0.465 & spa\_Latn & 0.421 & eng\_Latn & \textbf{0.484} & arb\_Arab \\
arz\_Arab & 0.325 & \textbf{0.430} & arb\_Arab & \textbf{0.430} & arb\_Arab & \textbf{0.430} & arb\_Arab & 0.325 & eng\_Latn & \textbf{0.430} & arb\_Arab \\
asm\_Beng & 0.574 & 0.548 & arb\_Arab & \textbf{0.600} & hin\_Deva & \textbf{0.600} & hin\_Deva & \textbf{0.600} & hin\_Deva & \textbf{0.600} & hin\_Deva \\
\bottomrule
\end{tabular}
    }
    \caption{Cross-Lingual Transfer Results of Taxi1500 (Part 1): The first column is the target language. For each language similarity measure, we report both the source language selected based on similarity and also the evaluation results on target language using the source language. For \mPLMMeasure, we report the layer achieving best performance (layer 4).}
    \label{tab:taxi1}
\end{table*}

\begin{table*}
    \centering
    \resizebox{0.925\textwidth}{!}{
\begin{tabular}{c|c|cc|cc|cc|cc|cc}
    \toprule
& ENG & \multicolumn{2}{c|}{LEX} & \multicolumn{2}{c|}{GEN} & \multicolumn{2}{c|}{GEO} & \multicolumn{2}{c|}{FEA} & \multicolumn{2}{c}{mPLM-Sim} \\
\midrule
hye\_Armn & 0.650 & \textbf{0.678} & arb\_Arab & 0.650 & eng\_Latn & \textbf{0.678} & arb\_Arab & 0.654 & rus\_Cyrl & 0.654 & rus\_Cyrl \\
ibo\_Latn & 0.544 & 0.544 & eng\_Latn & 0.544 & eng\_Latn & \textbf{0.566} & spa\_Latn & 0.544 & eng\_Latn & \textbf{0.566} & spa\_Latn \\
ilo\_Latn & 0.511 & 0.511 & eng\_Latn & 0.511 & eng\_Latn & 0.463 & cmn\_Hani & 0.511 & eng\_Latn & \textbf{0.591} & spa\_Latn \\
ind\_Latn & 0.720 & 0.720 & eng\_Latn & 0.720 & eng\_Latn & \textbf{0.795} & hin\_Deva & 0.720 & eng\_Latn & 0.720 & eng\_Latn \\
isl\_Latn & 0.497 & 0.497 & eng\_Latn & 0.497 & eng\_Latn & 0.497 & eng\_Latn & 0.497 & eng\_Latn & \textbf{0.602} & spa\_Latn \\
ita\_Latn & \textbf{0.608} & 0.593 & spa\_Latn & 0.593 & spa\_Latn & 0.593 & spa\_Latn & 0.593 & spa\_Latn & 0.593 & spa\_Latn \\
jav\_Latn & 0.445 & 0.445 & eng\_Latn & 0.445 & eng\_Latn & 0.428 & cmn\_Hani & 0.441 & hin\_Deva & \textbf{0.516} & spa\_Latn \\
\bottomrule
\end{tabular}
    }
    \caption{Cross-Lingual Transfer Results of Taxi1500 (Part 2): The first column is the target language. For each language similarity measure, we report both the source language selected based on similarity and also the evaluation results on target language using the source language. For \mPLMMeasure, we report the layer achieving best performance (layer 4).}
    \label{tab:taxi2}
\end{table*}

\begin{table*}
    \centering
    \resizebox{0.925\textwidth}{!}{
\begin{tabular}{c|c|cc|cc|cc|cc|cc}
    \toprule
& ENG & \multicolumn{2}{c|}{LEX} & \multicolumn{2}{c|}{GEN} & \multicolumn{2}{c|}{GEO} & \multicolumn{2}{c|}{FEA} & \multicolumn{2}{c}{mPLM-Sim} \\
\midrule
plt\_Latn & 0.503 & 0.503 & eng\_Latn & 0.503 & eng\_Latn & 0.495 & arb\_Arab & \textbf{0.627} & rus\_Cyrl & 0.562 & spa\_Latn \\
pol\_Latn & \textbf{0.690} & \textbf{0.690} & eng\_Latn & 0.690 & rus\_Cyrl & \textbf{0.690} & eng\_Latn & 0.690 & rus\_Cyrl & 0.690 & rus\_Cyrl \\
por\_Latn & \textbf{0.615} & 0.605 & spa\_Latn & 0.605 & spa\_Latn & 0.605 & spa\_Latn & \textbf{0.615} & eng\_Latn & 0.605 & spa\_Latn \\
prs\_Arab & 0.677 & 0.653 & arb\_Arab & 0.665 & hin\_Deva & 0.665 & hin\_Deva & \textbf{0.691} & cmn\_Hani & 0.665 & hin\_Deva \\
quy\_Latn & 0.696 & 0.696 & eng\_Latn & 0.696 & eng\_Latn & 0.693 & spa\_Latn & \textbf{0.718} & hin\_Deva & 0.693 & spa\_Latn \\
ron\_Latn & 0.582 & 0.582 & eng\_Latn & \textbf{0.617} & spa\_Latn & 0.582 & eng\_Latn & 0.589 & rus\_Cyrl & \textbf{0.617} & spa\_Latn \\
run\_Latn & 0.470 & 0.470 & eng\_Latn & 0.470 & eng\_Latn & 0.508 & arb\_Arab & \textbf{0.546} & hin\_Deva & 0.504 & spa\_Latn \\
\bottomrule
\end{tabular}
    }
    \caption{Cross-Lingual Transfer Results of Taxi1500 (Part 3). The first column is the target language. For each language similarity measure, we report both the source language selected based on similarity and also the evaluation results on target language using the source language. For \mPLMMeasure, we report the layer achieving best performance (layer 4).}
    \label{tab:taxi3}
\end{table*}

\end{document}